%% file: main.tex
\documentclass[10pt,twocolumn,letterpaper]{article}

\usepackage{cvpr}
\usepackage{times}
\usepackage{epsfig}
\usepackage{graphicx}
\usepackage{amsmath}
\usepackage{amssymb}
\usepackage{placeins}
\usepackage{color}
\usepackage{lipsum}
\usepackage{enumitem}
\usepackage{soul}

\usepackage{url}
\usepackage{mathtools}
\usepackage{subcaption}

\usepackage[export]{adjustbox}
\usepackage{makecell}
\usepackage{siunitx}

\usepackage[utf8]{inputenc} 
\usepackage[T1]{fontenc}    
\usepackage{url}            
\usepackage{booktabs}       
\usepackage{amsfonts}       
\usepackage{nicefrac}       
\usepackage{microtype}      
\usepackage{caption}
\usepackage{epsfig} 
\usepackage{graphicx} 
\usepackage{dsfont}
\usepackage{amsmath,amssymb} 
\usepackage{color}

\usepackage{bbm} 

\usepackage[pagebackref=true,breaklinks=true,letterpaper=true,colorlinks,bookmarks=false]{hyperref}

\usepackage[ruled]{algorithm}
\usepackage{tabularx}
\usepackage{multirow}
\usepackage{xcolor}

\usepackage{color}
\definecolor{gray}{rgb}{0.5,0.5,0.5}

\newcommand{\ignore}[1]{}

\iftrue  
        \newcommand{\todo}[1]{}
        \newcommand{\outline}[1]{}
        \newcommand{\textgray}[1]{}
        \newcommand{\commenttext}[1]{}
        \newcommand{\commentfoot}[1]{}
        \newcommand{\commentselfoot}[2]{}
        \newcommand{\topic}[1]{}
\else 
        \newcommand{\todo}[1]{{\textcolor{red}{[[TODO: {#1}]]}}}
        \newcommand{\outline}[1]{{\textcolor{blue}{[[{#1}]]}}}
        \newcommand{\textgray}[1]{\textcolor{gray}{[[{#1}]]}}
        \newcommand{\commenttext}[1]{\textcolor{red}{[[{#1}]]}}
        \newcommand{\commentfoot}[1]{\footnote{\textcolor{red}{\textit{#1}}}}
        \newcommand{\commentselfoot}[2]{{\textcolor{blue}{#1}}\commenttext{#2}}
        \newcommand{\topic}[1]{\textcolor{gray}{\textbf{(#1.)}}}
\fi

\newcommand{\duygu}[1]{\textcolor{blue}{[duygu: {#1}]}}

\iffalse

\else 

\fi

\newcommand{\lblfig}[1]{\label{fig:#1}}

\newcommand{\ignorethis}[1]{}

\newcommand\blfootnote[1]{%
  \begingroup
  \renewcommand\thefootnote{}\footnote{#1}%
  \addtocounter{footnote}{-1}%
  \endgroup
}

\cvprfinalcopy 

\def\cvprPaperID{1988} 

\ifcvprfinal\fi
\begin{document}

\title{Neural Kinematic Networks for Unsupervised Motion Retargetting}

\author{Ruben Villegas \textsuperscript{\textnormal{1,*}} 
\qquad Jimei Yang \textsuperscript{\textnormal{2}}
\qquad Duygu Ceylan \textsuperscript{\textnormal{2}}
\qquad Honglak Lee \textsuperscript{\textnormal{1,3}} \\
\textsuperscript{\textnormal{1}}University of Michigan, Ann Arbor \\
\textsuperscript{\textnormal{2}}Adobe Research \\
\textsuperscript{\textnormal{3}} Google Brain \\
}

\twocolumn[{%
\renewcommand\twocolumn[1][]{#1}%
\vspace{-3em}
\maketitle
\vspace{-3em}
\begin{center}
    \centering
    \includegraphics[width=0.9\linewidth]{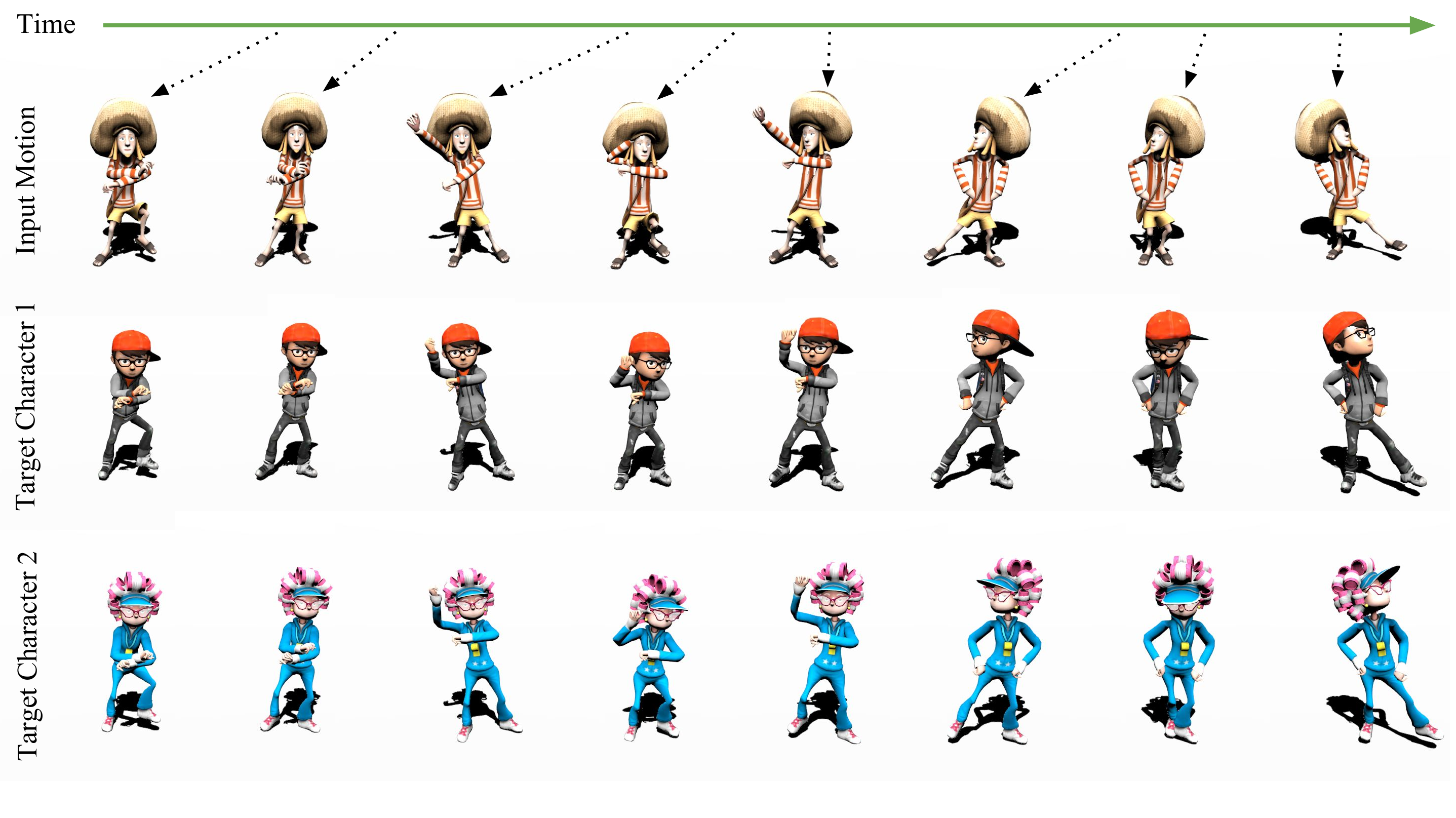}
    \vspace{-20pt}
    \captionof{figure}{Our end-to-end method retargets a given input motion (top row), to new characters with different bone lengths and proportions, (middle and bottom row). The target characters are never seen performing the input motion during training.}
    \lblfig{teaser}
\end{center}%
}]

\maketitle
\thispagestyle{empty}

\blfootnote{* Most of this work was done during Ruben's internship at Adobe.}

\input{0_abstract.tex}

\input{1_introduction.tex}

\input{2_relatedwork.tex}

\input{3_background.tex}
\input{4_method.tex}

\input{5_experiments.tex}

\input{6_conclusion.tex}

{\small
\bibliographystyle{ieee}
\bibliography{egbib}
}

\input{7_supp.tex}

\end{document}

%% file: 0_abstract.tex
\vspace{-12pt}
\begin{abstract}
\vspace{4pt}
We propose a recurrent neural network architecture with a Forward Kinematics layer and cycle consistency based adversarial training objective for unsupervised motion retargetting. Our network captures the high-level properties of an input motion by the forward kinematics layer, and adapts them to a target character with different skeleton bone lengths (e.g., shorter, longer arms etc.). Collecting paired motion training sequences from different characters is expensive. Instead, our network utilizes cycle consistency to learn to solve the Inverse Kinematics problem in an unsupervised manner. Our method works \emph{online}, i.e., it adapts the motion sequence on-the-fly as new frames are received. In our experiments, we use the Mixamo animation data~\footnote{\url{https://www.mixamo.com}. See details in Section 5.} to test our method for a variety of motions and characters and achieve state-of-the-art results. We also demonstrate motion retargetting from monocular human videos to 3D characters using an off-the-shelf 3D pose estimator.
\end{abstract}

%% file: 1_introduction.tex
\vspace{-20pt}
\section{Introduction}
\label{sec:intro}
\vspace{-3pt}

Imitation is an important learning scheme for agents to acquire motor control skills~\cite{schaal1999imitation}. 
It is often formulated as learning from expert demonstrations with access to sample trajectories of state-action pairs~\cite{Bagnell,ho2016generative}.
However, this first-person imitation assumption may not always hold since 1) the teacher and the learner may have different physical structures, e.g., a human being vs a humanoid robot~\cite{bin2015kinodynamically,TCN2017} and 2) the learner may only observe the states of the teacher, e.g. joint positions, but not the actions that generate these states~\cite{merel2017learning}. 
Adapting the motion of the teacher, e.g., a person, to the learner, e.g., a humanoid robot~\cite{Ayusawa2017TRO} or an avatar~\cite{Shon2006Imitation, mehta2017vnect}, is often referred as motion retargetting in robotics and computer animation.
This paper focuses on retargetting motions from a source to any target character with a known but different kinematic structure in terms of bone lengths and proportions.
Skeletal differences between the source and target characters create the necessity of disentangling skeleton-independent features of the source motion and automatically adapting them to a target character in one shot, ideally without any post-processing optimization and hand-tuning steps. Furthermore, a faithful solution needs to ensure the retargetted motion to be natural and realistic-looking which has been a long-standing challenge for animation.

Deep neural networks are known to have the ability to learn high-level features in sequential data that humans may not be able to easily identify, and have already achieved remarkable performance in machine translation~\cite{johnson2016google} and speech recognition~\cite{graves2013speech}. 
However, human motions are highly nonlinear and intrinsically constrained by kinematic structures of the skeletons. Thus classic sequence models such as recurrent neural networks (RNNs) may not be directly applicable to motion retargetting.

In this paper, we propose a novel neural network architecture to perform motion retargetting between characters with different skeleton structures (i.e., same topology but different bone length proportions).
Our architecture relies on an analytic \textit{Forward Kinematics} layer and two RNNs that work together to (i) encode the input motion data to motion features, and (ii) decode the joint rotations of the target skeleton from the identified features. The forward kinematics layer takes as input the joint rotations and the T-pose of a target skeleton, and renders the resulting motion. This fully differentiable layer forces the network to discover valid joint rotations by enabling to reason about the realism of the resulting motion. 
We use an adversarial training objective, rooted on the cycle consistency principle~\cite{CycleGAN2017}, to learn motion retargetting in an unsupervised way.
In particular, the motion retargetted onto a target character should generate the original motion of the source character when retargetted back. Furthermore, the generated motion should be as natural as other known motions of the target character for an adversarially trained discriminator. The decoder RNN is conditioned on the target character, and together with the adverserial training, is able to generate natural motions for unseen characters as well. In our experiments, we show that the proposed method can perform \emph{online} motion retargetting, i.e., adapting the input motion sequence on-the-fly as new frames are received. We also use 3D pose estimates from video sequences, e.g., in Human 3.6M dataset~\cite{h36m_pami}, as input to our network to animate Mixamo 3D characters.

The contributions of our work are summarized below:
\begin{itemize}
\item A novel \textit{Neural Kinematic Network} consisting of two RNNs and a forward kinematics layer that automatically discovers the necessary joint rotations (i.e., solution to the \textit{Inverse Kinematics (IK)} problem) for motion retargetting without requiring ground-truth rotations during training.
\item A sequence-level adversarial cycle consistency objective function for unsupervised learning for motion retargetting which does not require input/output motion pairs of different skeletons during training.
\end{itemize}

\ignore{
\duygu{I usually don't have such an organization paragraph since the structure of a paper is more or less clear.}
The paper is organized as follows:
We review related work in Section~\ref{sec:related}, and give some background about the problem at hand in Section~\ref{sec:background}.
Our method for unsupervised motion retargetting is described in Section~\ref{sec:method}, and detailed experimental results are illustrated in Section~\ref{sec:experiments}.
Finally, we discuss our method, and talk about future work in Section~\ref{sec:conclusion}.
}

%% file: 2_relatedwork.tex
\section{Related work}
\label{sec:related}
%
Gleicher~\cite{gleicher} first formulated motion retargetting as a spacetime optimization problem with kinematic constraints that is solved for the entire motion sequence.
Lee and Shin~\cite{lee1999hierarchical} proposed a decomposition approach that first solves the IK problem for each frame to satisfy the constraints and then fits multilevel B-spline curves to achieve smooth results. 
Tak and Ko~\cite{tak2005physically} further added dynamics constraints to perform sequential filtering to render physically plausible motions.
%
%
Choi and Ko~\cite{choi} proposed an online retargetting method by solving per-frame IK that computes the change in joint angles corresponding to the change in end-effector positions while imposing motion similarity as a secondary task. 
While the above-mentioned approaches require iterative optimization with hand-designed kinematic constraints for particular motions, our method learns to produce proper and smooth changes of joint angles (solving IK) in one-pass feed-forward inference of RNNs, and is able to generalize to unseen characters and novel motions. 
The idea of solving approximate IK can be traced back to the early blending-based methods~\cite{rose2001artist,kovar2004automated}.
A target skeleton can be viewed as a new style. Our method can be applied to motion style transfer that has been a popular research area in computer animation~\cite{brand2000style,hsu2005style,min2010synthesis,xia2015realtime,yumer2016spectral}.

%

Different machine learning algorithms have been used in modeling human motions. Early works used auto-regressive RBMs~\cite{taylor2007modeling} or Gaussian process dynamic models~\cite{wang2008gaussian, grochow2004style} to learn human motions in small scale. 
In particular, Grochow et al.~\cite{grochow2004style} solves IK by constraining the generated poses to a learned Gaussian process prior.
With the surge of deep learning, a variety of neural networks have been used to synthesize human motions~\cite{fragkiadaki2015recurrent,Holden,jain2016structural,butepage2017deep,martinez2017human,li2017auto}.
These networks are not applicable to motion retargetting as they directly generate the xyz-coordinates of joints and thus require a further post-processing to ensure bone length consistency.
Instead, our method predicts quaternions that represent the rotation of each joint with respect to the T-pose without rotation supervision, which admits an end-to-end solution to motion retargetting and also has the potential of synthesizing kinematically plausible motions.
Notably, Jain \etal~\cite{jain2016structural} model human motions with a spatial-temporal graph that considers the skeletal structure but not in an analytic form.

Our work is also related to research efforts on ``vision as inverse graphics''. Differentiable rendering layers are incorporated into deep neural networks to disentangle imaging factors of rigid objects, such as 3D shape, camera, normal map, lighting and materials~\cite{yan2016perspective,rezende2016unsupervised,tulsiani2017multi,liu2017material}. Wu et al.~\cite{vda} further incorporated a differentiable physics simulator~\cite{chang2016compositional} to disentangle physical properties of multiple rigid objects. Our network disentangles the hierarchical rotations of articulated skeletons through a differentiable forward kinematics layer. 

%% file: 3_background.tex
\section{Background}
\label{sec:background}
We first introduce some concepts in robotics and computer animation essential for building our model.
\subsection{Forward kinematics}
\label{sec:fk}
\begin{figure}[t]
    \begin{center}
	    \includegraphics[width=0.95\linewidth]{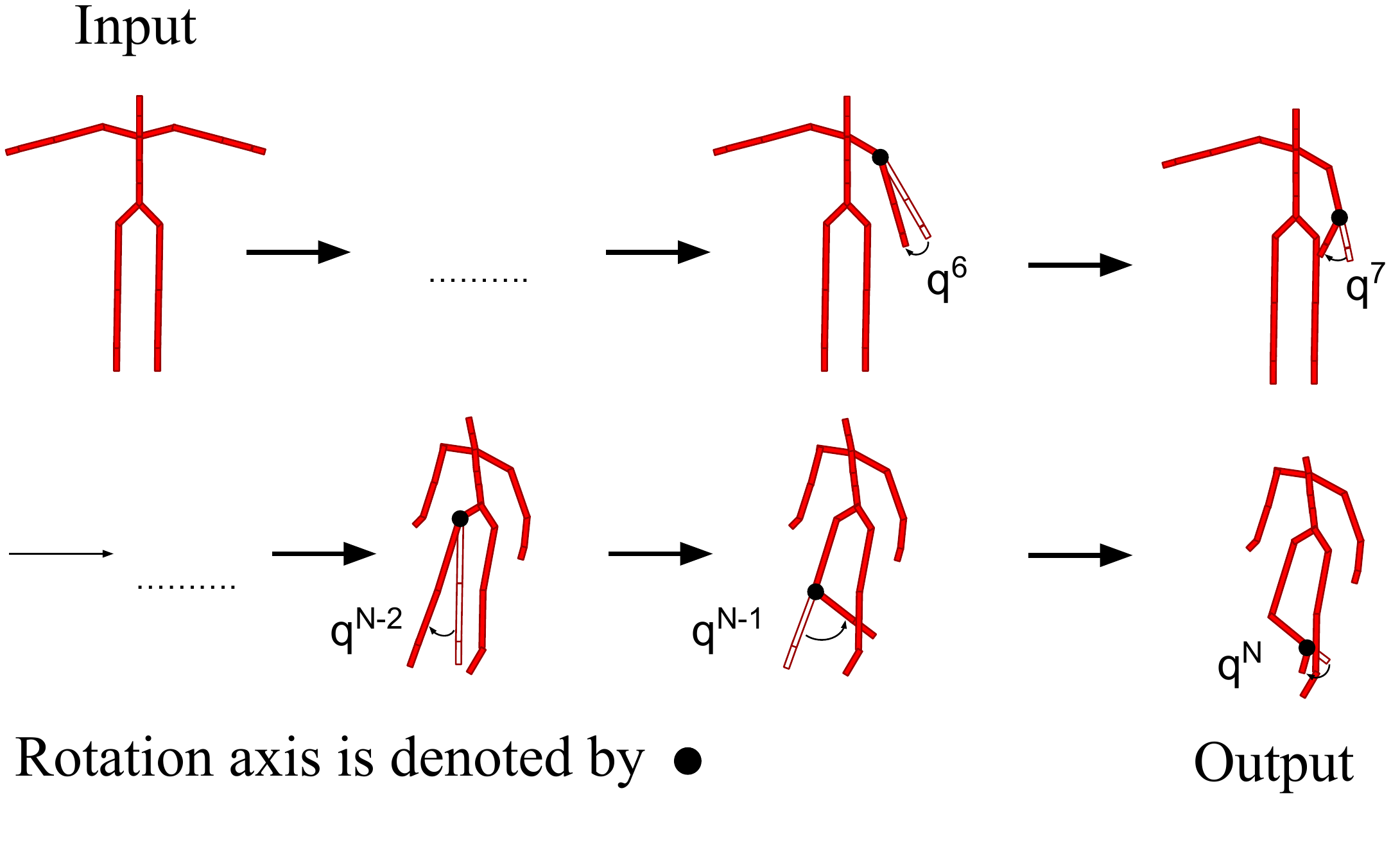}
	\end{center}
	\vspace{-20pt}
	\caption{Forward kinematics from T-pose skeleton. Starting from the input skeleton, the forward kinematics layer rotates bones to achieve the desired output configuration.}
	\vspace{-.1in}
\label{fig:fk}
\end{figure}

Forward kinematics (FK) refers to the process of computing the positions of skeleton joints, also known as $\textit{end-effectors}$, in 3D space given the joint rotations and initial positions.
FK is performed by recursively rotating the joints of an input skeleton tree starting from the root joint and ending in the leaf joints, and is defined by:
\begin{equation}
p^n=p^{parent(n)} + R^n \bar{s}^n, \nonumber
\label{eq:fk}
\end{equation}
where $p^n\in\mathbb{R}^3$ is the updated 3D position of the $n$-th joint and $p^{parent(n)}\in\mathbb{R}^3$ is the current position of its parent.
$R^n\in\mathbb{SO}(3)$ is the rotation of the $n$-th joint with respect to its parent. $\bar{s}^n\in\mathbb{R}^3$ is the 3D offset of the $n$-th joint relative to its parent in the input skeleton, and is defined by:
\begin{equation}
\bar{s}^n=\bar{p}^n-\bar{p}^{parent(n)}\nonumber,
\label{eq:sk}
\end{equation}
Note that $\bar{p}^n$ and $\bar{p}^{parent(n)}$ refer to joint positions in the input \emph{T-pose skeleton} as depicted in Figure \ref{fig:fk}.

\subsection{Inverse kinematics}
While FK refers to computing the 3D joint locations by recursively applying joint rotations, inverse kinematics (IK) is the reverse process of computing joint rotations $R^{1:N}$ that ensure specific joints are placed at the desired target locations $p^{1:N}$ starting from initial positions $p^{1:N}_0$. Thus, IK is defined by:
\begin{equation}
R^{1:N}=\text{IK}(p^{1:N}, p^{1:N}_0).\nonumber
\label{eq:ik}
\end{equation}
IK is inherently an ill-posed problem.
Target configuration of joint locations can be fulfilled by multiple joint rotations or no joint rotations. 
Classic IK solutions often resort to iterative optimization by calculating the inverse Jacobian of the highly nonlinear FK function numerically or analytically.
\ignore{
In motion retargetting methods tackle this problem by using iterative optimization techniques to minimize hand-designed objective functions that, for example, give more importance to certain end-effectors or focus on foot contacts with the floor~\cite{gleicher}.
In contrast, our method avoids the need of hand-designed objective functions by learning transferable motion features in one shot. Our network internally performs forward kinematics to disentangle motion features, i.e. joint rotations, from the joint locations.
}

%% file: 4_method.tex
\section{Method}
\label{sec:method}

In this section, we present our proposed method for unsupervised motion retargetting.
There are two main components: (i) the neural kinematic network architecture for skeleton conditioned motion synthesis, and (ii) the adversarial cycle consistency training for unsupervised motion retargetting. We next describe these components in detail.

\subsection{Neural kinematic networks} \label{sec:mo_synth}
Our neural kinematic networks for motion synthesis component is built to strictly manipulate a target skeleton, which we refer as \emph{condition skeleton}, into performing a given motion sequence performed by another source skeleton through a $\textit{Forward Kinematics}$ layer.

In our setup, the input motion data $x_{1:T}$ is decomposed into $p_{1:T}$ and $v_{1:T}$, where for each time $t$, $p_t\in\mathbb{R}^{3N}$ represents the local \textit{xyz}-configuration of the skeleton's pose with respect to its root joint (i.e., hip joint), and $v_t\in\mathbb{R}^4$ represents the global motion of the skeleton's root joint (i.e., \textit{x},\textit{y},\textit{z}-velocities and rotation with respect to the axis perpendicular to the ground).
Given the condition skeleton, the motion synthesis module outputs the rotations, $R_t^n$, that are then applied to each joint $n$ at time $t$, as well as the global motion parameters.

\subsubsection{Forward kinematics layer} \label{sec:fk_layer}
At the core of our neural kinematic networks for motion synthesis component lies the \emph{Forward Kinematics layer} (Figure \ref{fig:fk}) which is designed to take in 3D rotations for each joint $n$ at time $t$ parameterized by $\textit{unit quaternions}$ $q_t^n\in\mathbb{R}^4$, and apply them to a skeleton bone configuration $\bar{s}^n$.
A \textit{quaternion} extends a \textit{complex number} in the form $r +x\mathbbm{i} +y\mathbbm{j} +z\mathbbm{k}$ and is used to rotate objects in 3 dimensional space, where $r$, $x$, $y$, and $z$ are real numbers and $\mathbbm{i}$, $\mathbbm{j}$, $\mathbbm{k}$ are quaternion units.
The rotation matrix corresponding to an input quaternion is calculated as follows:

\begin{equation}
R^n_t {=} \left( \begin{smallmatrix}
1-2({q_{tj}^n}^2+{q_{tk}^n}^2) & 2(q_{ti}^n q_{tj}^n+q_{tk}^n q_{tr}^n) &  2(q_{ti}^n q_{tk}^n-q_{tj}^n q_{tr}^n)\\
2(q_{ti}^n q_{tj}^n-q_{tk}^n q_{tr}^n) & 1-2({q_{ti}^n}^2+{q_{tk}^n}^2) &  2(q_{tj}^n q_{tk}^n+q_{ti}^n q_{tr}^n)\\
2(q_{ti}^n q_{tk}^n+q_{tj}^n q_{tr}^n) & 2(q_{tj}^n q_{tk}^n-q_{ti}^n q_{tr}^n) & 1-2({q_{ti}^n}^2+{q_{tj}^n}^2)
\end{smallmatrix} \right)
\label{eq:rot_mat}
\end{equation}

Given the rotation matrices $R^n_t\in\mathbb{SO}(3)$ for each joint, the FK layer updates the joint positions of the condition skeleton by applying these rotations in a recursive manner as described in Section~\ref{sec:fk} and shown in Figure~\ref{fig:fk},
\begin{equation}
p^{1:N}_t = \text{FK}(q^{1:N}_t, \bar{s}).\nonumber
\label{eq:FK_layer}
\end{equation}

The FK layer serves as a tool for mapping the joint rotations to actual joint locations and thus helps our network to focus on learning skeleton independent motion features, i.e., joint rotations. 

\subsubsection{Online motion synthesis} \label{sec:online}

\begin{figure}[t]
    \begin{center}
	    \includegraphics[width=1.0\linewidth]{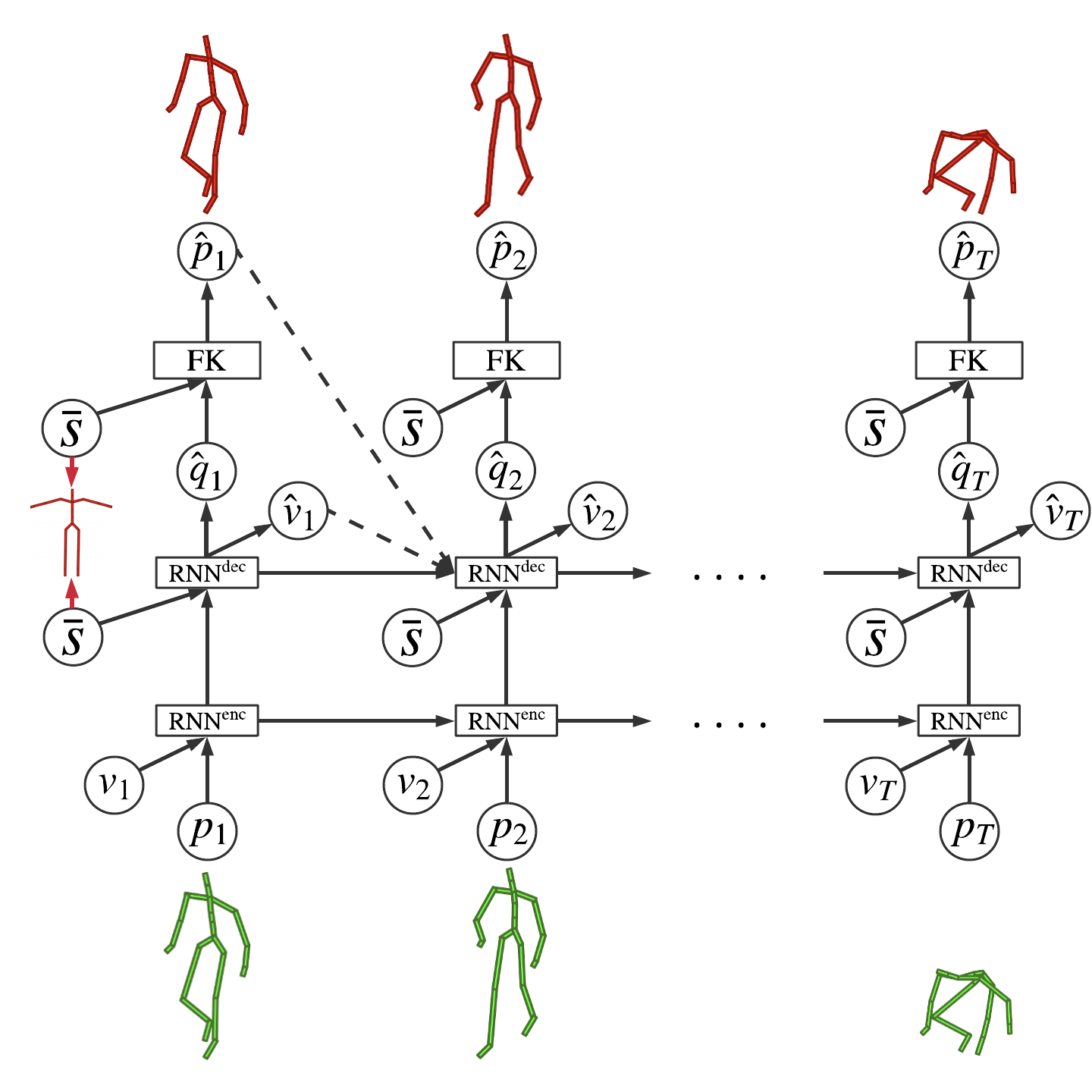}
	\end{center}
	\vspace{-22pt}
	\caption{Neural kinematic networks for motion synthesis.}
	\vspace{-.1in}
\label{fig:online}
\end{figure}

Our proposed neural kinematic networks architecture for online motion synthesis is shown in Figure \ref{fig:online}. Taking advantage of the temporal coherency in motion sequences, we synthesize the current motion step at time $t$ by conditioning on previous steps through an RNN hidden representation. 
\ignore{
Specifically, the current step in the input motion is encoded by
\begin{equation}
h_t^{\text{enc}} = \text{RNN}^\text{enc}(x_t^A, h_{t-1}^{\text{enc}}),
\label{eq:online_enc}
\end{equation}
where $\text{RNN}^\text{enc}(.,.)$ is an encoder RNN, $h_t^{\text{enc}}$ is the encoding of the input motion up to time $t$, and $x_t^A=[p_t^A, v_t^A]$ is the current input.
The encoded feature is then fed to a decoder RNN to perform skeleton conditioned motion synthesis by:
\begin{align}
h_t^{\text{dec}} &= \text{RNN}^\text{dec}(\hat{x}_{t-1}^B, h_t^{\text{enc}}, \bar{s}^B, h_{t-1}^{\text{dec}}),\\
\hat{q}^B_t &= \frac{{W^p}^T h_t^{\text{dec}}}{\|{W^p}^T h_t^{\text{dec}}\|},\\
\hat{p}^B_t \ &= \text{FK}(\hat{q}^B_t, \bar{s}^B),\\
\hat{v}^B_t \ &= {W^v}^T h_t^{\text{dec}},\\
\hat{x}^B_t \ &= \left[\hat{p}^B_t,\hat{v}^B_t\right].
\end{align}
where $h_t^{\text{dec}}$ is the encoding of the motion performed by both, skeleton $A$ and skeleton $B$, $\hat{x}^B_t$ is the synthesized motion at time step $t$ for the target skeleton $\bar{s}^B$. 
The unit vector $\hat{q}^B_t\in\mathbb{R}^{4N}$ denotes the rotations --- which can be interpreted as actions --- to be applied to the condition skeleton through the FK layer.
The outputs $\hat{p}^B_t$ and $\hat{v}^B_t$ are the estimated local and global motion of the condition skeleton $B$.
Finally, $W^{v}\in\mathbb{R}^{d \times 4}$, and $W^{p}\in\mathbb{R}^{d \times 4N)}$ are learnable parameters.
}

The current step in the input motion is encoded by:
\begin{equation}
h_t^{\text{enc}} = \text{RNN}^\text{enc}(x_t, h_{t-1}^{\text{enc}}; W^{enc}),
\label{eq:online_enc}
\end{equation}
where $\text{RNN}^\text{enc}(.,.)$ is an encoder RNN, $h_t^{\text{enc}}$ is the encoding of the input motion up to time $t$, and $x_t=[p_t, v_t]$ is the current input.
The encoded feature is then fed to a decoder RNN to perform skeleton conditioned motion synthesis by:
\begin{align}
h_t^{\text{dec}} &= \text{RNN}^\text{dec}(\hat{x}_{t-1}, h_t^{\text{enc}}, \bar{s}, h_{t-1}^{\text{dec}}; W^{dec}),\\
\hat{q}_t &= \frac{{W^p}^T h_t^{\text{dec}}}{\|{W^p}^T h_t^{\text{dec}}\|},\\
\hat{p}_t \ &= \text{FK}(\hat{q}_t, \bar{s}),\\
\hat{v}_t \ &= {W^v}^T h_t^{\text{dec}},\\
\hat{x}_t \ &= \left[\hat{p}_t,\hat{v}_t\right].
\end{align}
where $h_t^{\text{dec}}$ is the hidden representation of decoder RNN, $\hat{x}_t$ is the synthesized motion at time step $t$ for the condition skeleton $\bar{s}$. 
The unit vector $\hat{q}_t\in\mathbb{R}^{4N}$ denotes the rotations --- which can be interpreted as actions --- to be applied to the condition skeleton through the FK layer.
The outputs $\hat{p}_t$ and $\hat{v}_t$ are the estimated local and global motion of the condition skeleton.
Finally, $W^{enc}$, $W^{dec}$, $W^{v}\in\mathbb{R}^{d \times 4}$ and $W^{p}\in\mathbb{R}^{d \times 4N}$ are learnable parameters.

When the condition skeleton is different from the skeleton where the input motion lives, the decoder is meant to generate the rotations of a new character to achieve motion retargetting.
Please note that in the rest of the paper, we use superscripts $A$ or $B$ to refer to the identity of the skeleton we are retargetting motion from and into.

\subsection{Adversarial cycle training for unsupervised motion retargetting} \label{sec:training}

\begin{figure}[h!]
    \vspace{-.17in}
    \begin{center}
	    \includegraphics[width=1\linewidth]{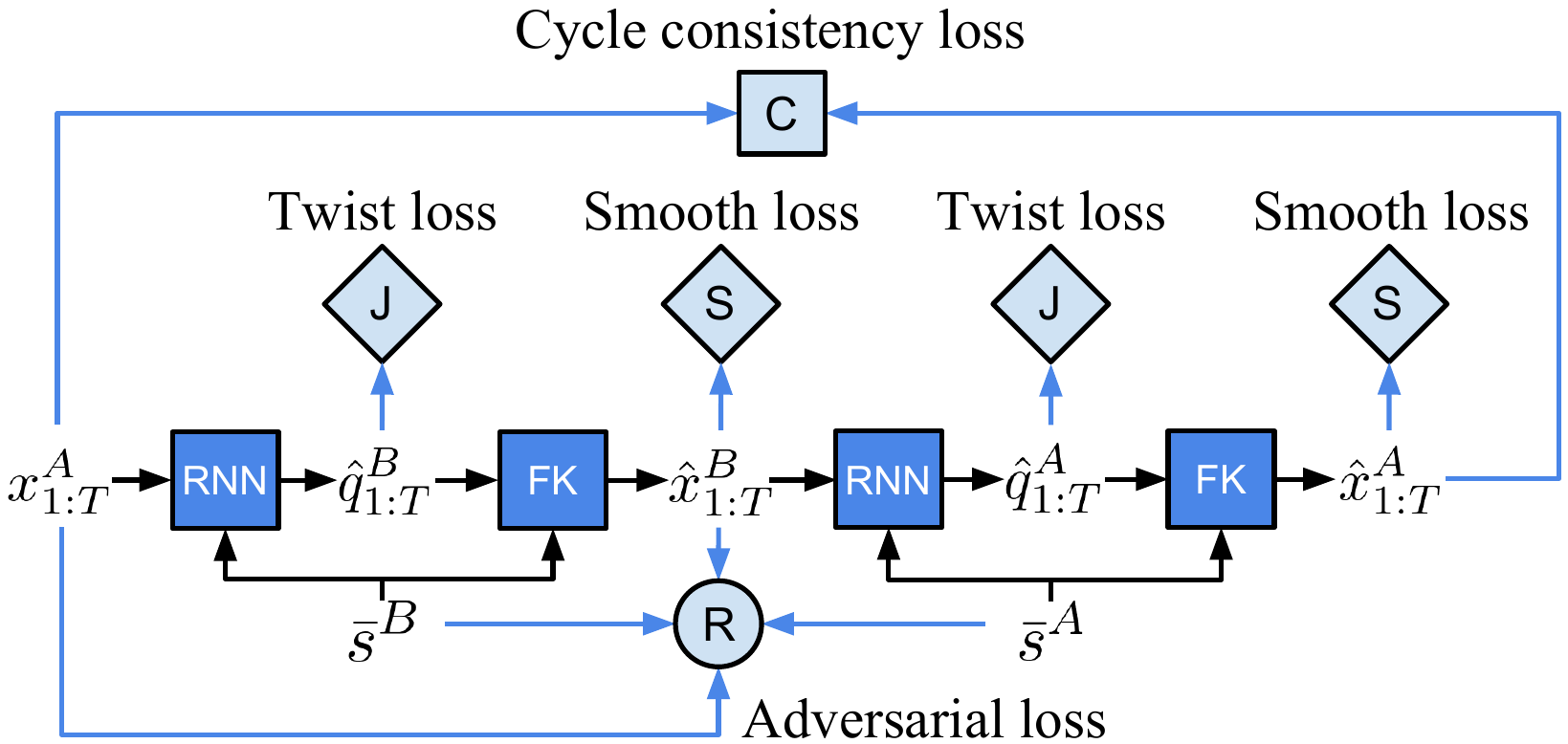}
	\end{center}
	\vspace{-12pt}
	\caption{Adversarial cycle consistency framework.}
	\vspace{-.1in}
\label{fig:training}
\end{figure}


In Section \ref{sec:mo_synth}, we describe a method for skeleton conditioned motion synthesis based on a forward kinematics layer embedded within the network architecture.
However, training such a network for motion retargetting is challenging as it is very expensive to collect paired motion data $x_t^A$ and $x_t^B$ where the same motion is performed by two different skeletons.
Note that collecting such data requires using iterative optimization based IK methods in addition to human hand-tuning of the retargetted motion.

We propose a training paradigm based on the cycle consistency principle~\cite{zhou2016learning} and adversarial training~\cite{NIPS2014_5423} for unsupervised motion retargetting (Figure \ref{fig:training}).
Let $f$ be our neural kinematic network, and let the superscripts define skeleton identity.
Given an input motion sequence from skeleton $A$, we first retarget the input motion to skeleton $B$ and back to $A$ as follows:
\begin{align}
&\hat{x}_{1:T}^B=f(x_{1:T}^A,\bar{s}^B), \\
&\hat{x}_{1:T}^A=f(\hat{x}_{1:T}^B,\bar{s}^A),
\end{align}
where $\hat{x}_{1:T}^B$ and $\hat{x}_{1:T}^A$ are synthesized motions for skeletons $B$ and $A$, respectively.
Therefore, we define four loss terms: adversarial loss on $\hat{x}_{1:T}^B$, cycle consistency loss on $\hat{x}_{1:T}^A$, twist loss on rotations $\hat{q}_{1:T}^A$ and $\hat{q}_{1:T}^B$, and smoothing loss on $\hat{v}_t^A$ and $\hat{v}_t^B$, so our full training objective is defined by:
\begin{align}
\min_f \max_d \ \ \ \ &C(\hat{x}_{1:T}^A, x_{1:T}^A) + R(\hat{x}_{1:T}^B, x_{1:T}^A) + \nonumber \\[-5pt]
&\lambda \ J(\hat{q}^B_{1:T}, \hat{q}^A_{1:T}) + \omega \ S(\hat{v}_{1:T}^B, \hat{v}_{1:T}^A),
\label{eq:full}
\end{align}
where $C$ is the cycle consistency loss, $R$ the adversarial loss, $J$ the joint twist loss, and $S$ the velocity smoothing loss.
\vspace{-8pt}
\paragraph{Adversarial loss.} The input motion $x_{1:T}^A=\left[p^A_{1:T},v^A_{1:T}\right]$, the synthesized motion $\hat{x}_{1:T}^B=\left[\hat{p}^B_{1:T},\hat{v}^B_{1:T}\right]$, and their respective skeleton are fed to a discriminator network $g$ that computes a \emph{realism score} for \emph{real} and \emph{fake} motion sequences:
\begin{align}
&r^A=g(p_{2:T}^A-p_{1:T-1}^A, v^A_{1:T-1}, \bar{s}^A), \\
&r^B=g(\hat{p}_{2:T}^B-\hat{p}_{1:T-1}^B, \hat{v}^B_{1:T-1}, \bar{s}^B),
\end{align}
where $r^A$ is the output of the discriminator given real data, and $r^B$ is the output of the discriminator given the fake data (i.e., the motion retargetted by our network into skeleton $B$).
The inputs to the discriminator $p_{2:T}^A-p_{1:T-1}^A$ and $p_{2:T}^B-p_{1:T-1}^B$ are the local motion difference between two adjacent time steps, and $\bar{s}^A$ and $\bar{s}^B$ denote the input and target skeletons $A$ and $B$, respectively.
During training, we randomly sample $\bar{s}^B$ from all the available skeletons, thus, it is possible for skeleton $B$ to be the same as skeleton $A$.
In case skeleton $B$ is the same as skeleton $A$, $\hat{x}_{1:T}^B = \hat{x}_{1:T}^A$, we switch between adversarial and square loss as follows: 
\begin{equation}
R(\hat{x}_{1:T}^B, x_{1:T}^A)=
\begin{cases}
      \|\hat{x}_{1:T}^B - x_{1:T}^A\|^2_2, & \text{if}\ B=A \\
      \log r^A + \beta \log ( 1 - r^B ), & \text{otherwise}.
\end{cases},
\label{eq:adv}
\end{equation}
When $B$ and $A$ are not the same, we rely on the motion distributions learned by $g$ as a training signal.
By observing other motion sequences performed by skeleton $B$, the discriminator network learns to identify motion behaviors of skeleton $B$.
The generator (encoder and decoder RNNs) uses this as indirect guidance to learn how the motion should be retargetted to $B$ and thus fool the discriminator.
When applying the adversarial loss, we use a balancing term $\beta$ to regulate the strength of the discriminator signal when optimizing $f$ to fool $g$.
We use $\beta=0.001$ in our experiments.
\vspace{-18pt}
\paragraph{Cycle consistency loss.} The cycle consistency loss $C$ optimizes the following objective:
\begin{equation}
C(\hat{x}_{1:T}^A, x_{1:T}^A) = \|x_{1:T}^A-\hat{x}_{1:T}^A\|^2_2.
\label{loss:cycle}
\end{equation}
Equation~\ref{loss:cycle} encourages $f$ to be able to take its own retargetted motion and map it back to the original motion source effectively achieving cycle consistency.
\vspace{-7pt}
\paragraph{Twist loss.} By optimizing the first two terms in Equation~\ref{eq:full}, our network discovers the necessary rotations to move the input skeleton end-effectors to the required positions for motion retargetting.
However, this does not prevent potential excessive bone twisting since \textit{xyz}-coordinates can be perfectly predicted regardless of how many times we rotate a bone around its own axis.
Thus, the third term in our objective constrains the bone rotations around its own axis.
\begin{align}
J(\hat{q}^B_{1:T}, \hat{q}^A_{1:T})= & \|\max(\mathbf{0}, |euler_{y}(\hat{q}^B_{1:T})| - \alpha)\|^2_2 + \nonumber\\
& \|\max(\mathbf{0}, |euler_{y}(\hat{q}^A_{1:T})| - \alpha)\|^2_2,
\end{align}
where $euler_{y}(.)$ converts the quaternion outputs of our network into rotation angles around the standard \textit{xyz}-axes and the subscript $y$ means to select the rotation angle around the plane parallel to the bone (i.e. \textit{y}-axis).
Therefore, any bone rotation exceeding $\alpha$ degrees in either negative or positive direction is penalized in our objective function.
We use $\alpha=100^\circ$, and $\lambda=10$ in our experiments.
\vspace{-8pt}
\paragraph{Smoothing loss.} Finally, the first two terms in our objective function treat global motion at each time step independently.
However, global motion in consecutive timesteps are highly dependent on each other, that is, global motion in the next timestep should change only slightly with respect to the previous global motion.
We constraint the global motion by:
\begin{align}
S(\hat{v}_{1:T}^B, \hat{v}_{1:T}^A)= & \| \hat{v}_{2:T}^B - \hat{v}_{1:T-1}^B \|^2_2 + \nonumber\\
& \| \hat{v}_{2:T}^A - \hat{v}_{1:T-1}^A \|^2_2,
\end{align}
We use $\omega=0.01$ in our experiments.

%% file: 5_experiments.tex
\section{Experiments}
\label{sec:experiments}

\textbf{Dataset.} We evaluate our method on the Mixamo dataset~\cite{Mixamo} which contains approximately 2400 unique motion sequences for 71 characters (i.e., skeletons).
For training, we collected non-overlapping motion sequences for 7 characters (AJ, Big Vegas, Goblin Shareyko, Kaya, Malcolm, Peasant Man, and Warrok Kurniawan) which in total results in 1646 training sequences at 30 frames per second.

For testing, we collected motion sequences for 6 characters (Malcolm, Mutant, Warrok Kurniawan, Sporty Granny, Claire, and Liam) and perform retargetting in four scenarios:
\begin{itemize}[noitemsep]
\vspace{-2pt}
\item Input motion is seen during training, and the target character is also seen during training but the target motion sequence is not.
\item Input motion is seen during training but the target character is never seen during training.
\item Input motion is not seen during training but the target character is seen during training.
\item Neither the input motion nor the target character are seen during training.
\vspace{-2pt}
\end{itemize}
Note that we also collected the ground truth retargetted motions of testing sequences for quantitative evaluation purposes only.
While we discuss our main findings below, detailed results and analysis of each scenario and character can be found in the supplementary material as well as details of how to acquire the exact training and testing data.

\textbf{Data preprocessing.} Each motion sequence is pre-processed by separating into local and global motion, similar to \cite{Holden}.
For local motion, we remove the global displacement (i.e., the motion of the root joint), and rotation around the axis vertical to the ground.
Global motion consists of the velocity of the root in the \textit{x}, \textit{y}, and \textit{z} directions, and an additional value representing the rotation around the axis perpendicular to the ground.
For training, and testing we use the following 22 joints: Root, Spine, Spine1, Spine2, Neck, Head, LeftUpLeg, LeftLeg, LeftFoot, LeftToeBase, RightUpLeg, RightLeg, RightFoot, RightToeBase, LeftShoulder, LeftArm, LeftForeArm, LeftHand, RightShoulder, RightArm, RightForeArm, and RightHand. 

\textbf{Baseline methods.} While there have been several optimization based approaches for the IK problem, most of these expect the user to provide motion specific constraints or goals. Since this is not feasible to do at a large scale, we instead show comparisons to learning based baseline methods that aim to identify such constraints automatically. The first baseline is an RNN architecture without the FK layer that directly outputs \textit{xyz}-coordinates for the local motion, and the global motion output is the same as ours.
Second, we use an MLP architecture that lacks recurrent connections, and directly outputs the \textit{xyz}-coordinates for the local motion, and the same global motion output as our method.
We also train both baselines with our adversarial cycle consistency objective.
Finally, we include another baseline that directly copies the per-joint rotation and the global motion of the input motion into the target skeleton.

\textbf{Training and evaluation.} We train our method and baselines by randomly sampling 2-second motion clips (60 frames) from the training sequences, and testing on motion clips of 4 seconds (120 frames) from the test sequences.
We initialized the quaternion outputs of the decoder RNN to be close to the identity rotation (i.e., close to zero rotation).
For training the discriminator network, we sample random motion sequences being performed by the same skeleton into which the motion synthesis network is retargetting motion.
Details of the network architecture and hyperparameters can be found in the supplementary material.
We perform two types of evaluations:
1) We evaluate the overall quality of the motion retargetting using a target character normalized Mean Square Error (MSE) on the estimated joint locations through time (i.e., \textit{xyz}-coordinates after combining local and global motion together).
2) We compare end-effector locations through time against the ground-truth.
3) We show qualitative results by rendering the animated 3D characters using the outputs of our network.

\subsection{Online Motion Retargetting From Character}

\begin{figure*}[htp!]
    \vspace{-.13in}
    \begin{center}
    \begin{subfigure}{0.95\linewidth}
        \hspace{-9pt}
	    \includegraphics[width=1.0\linewidth]{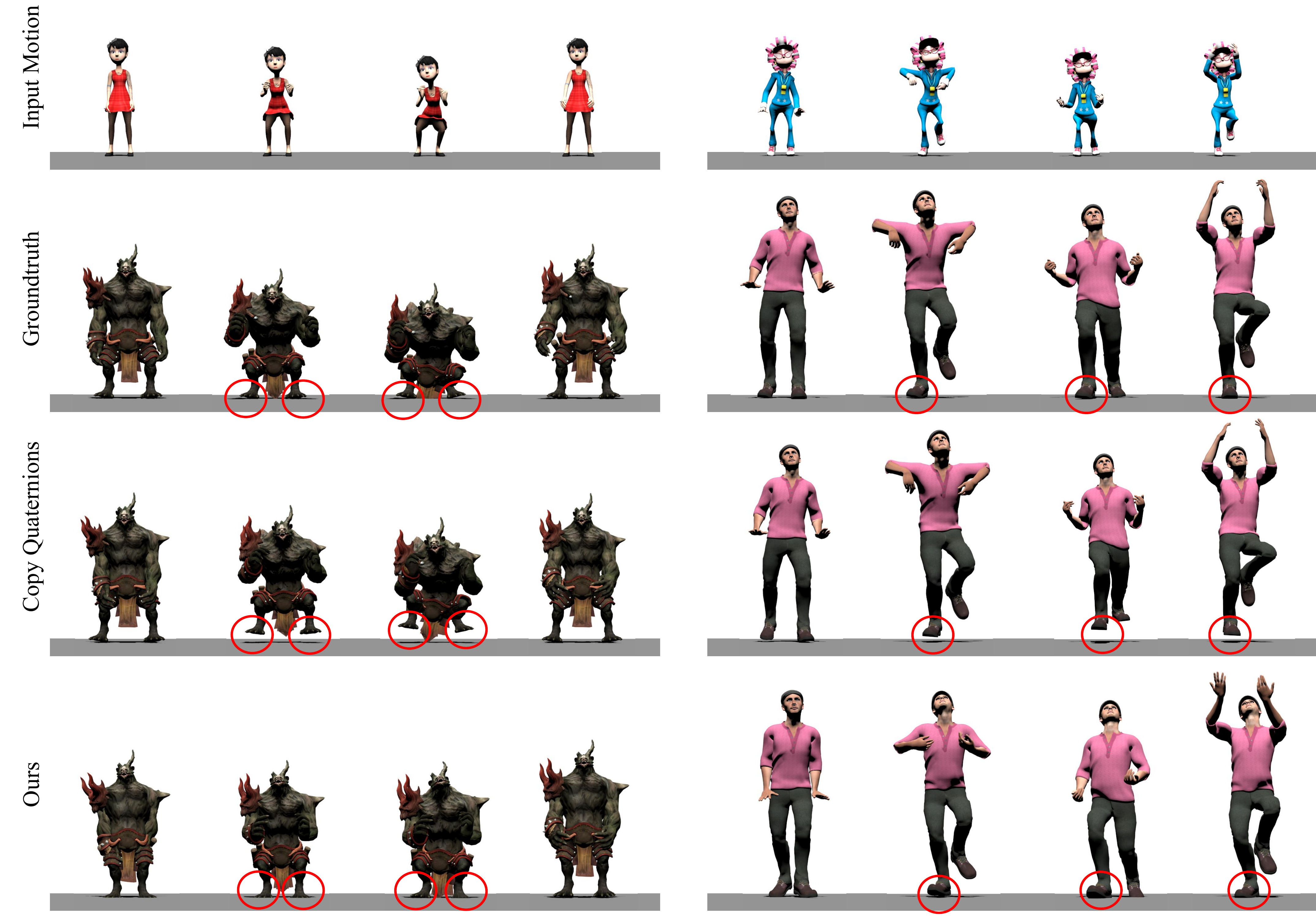}
	    \vspace{5pt}
	\end{subfigure}
	
	\begin{subfigure}{0.25\linewidth}
	\includegraphics[width=1.0\linewidth]{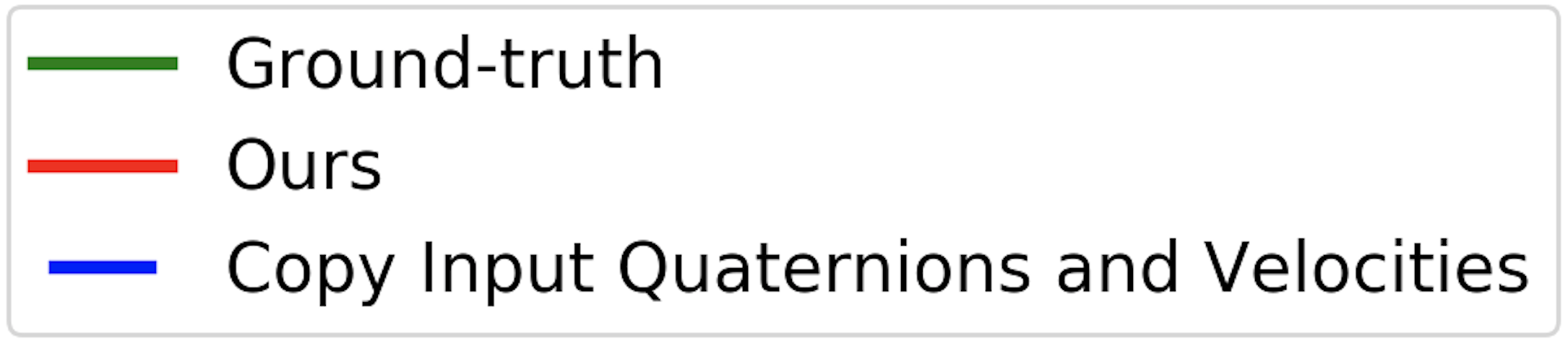}
	\vspace{-7pt}
	\end{subfigure}
	\begin{subfigure}{0.95\linewidth}
	    \includegraphics[width=.25\linewidth]{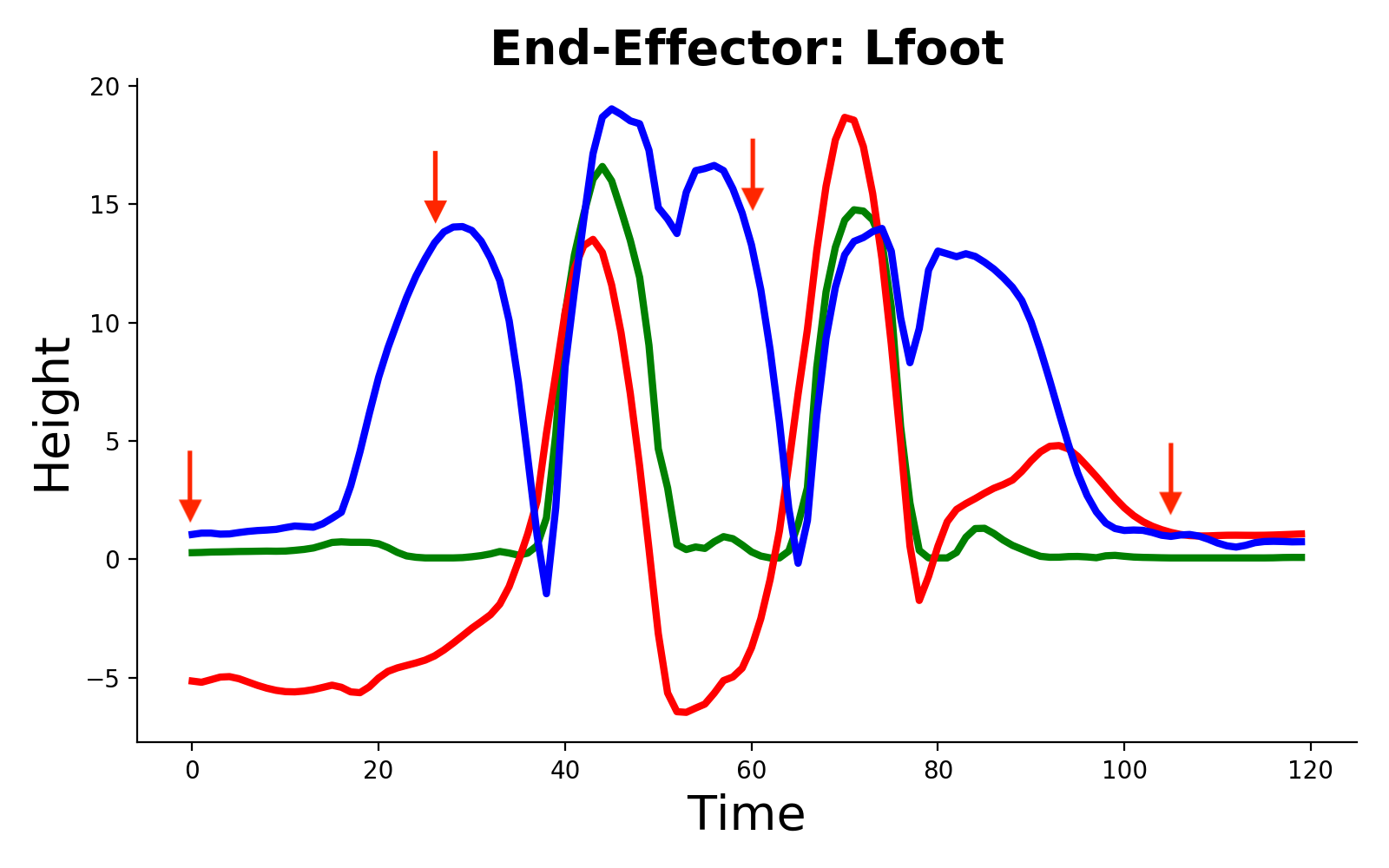} \hspace{-9pt}
	    \includegraphics[width=.25\linewidth]{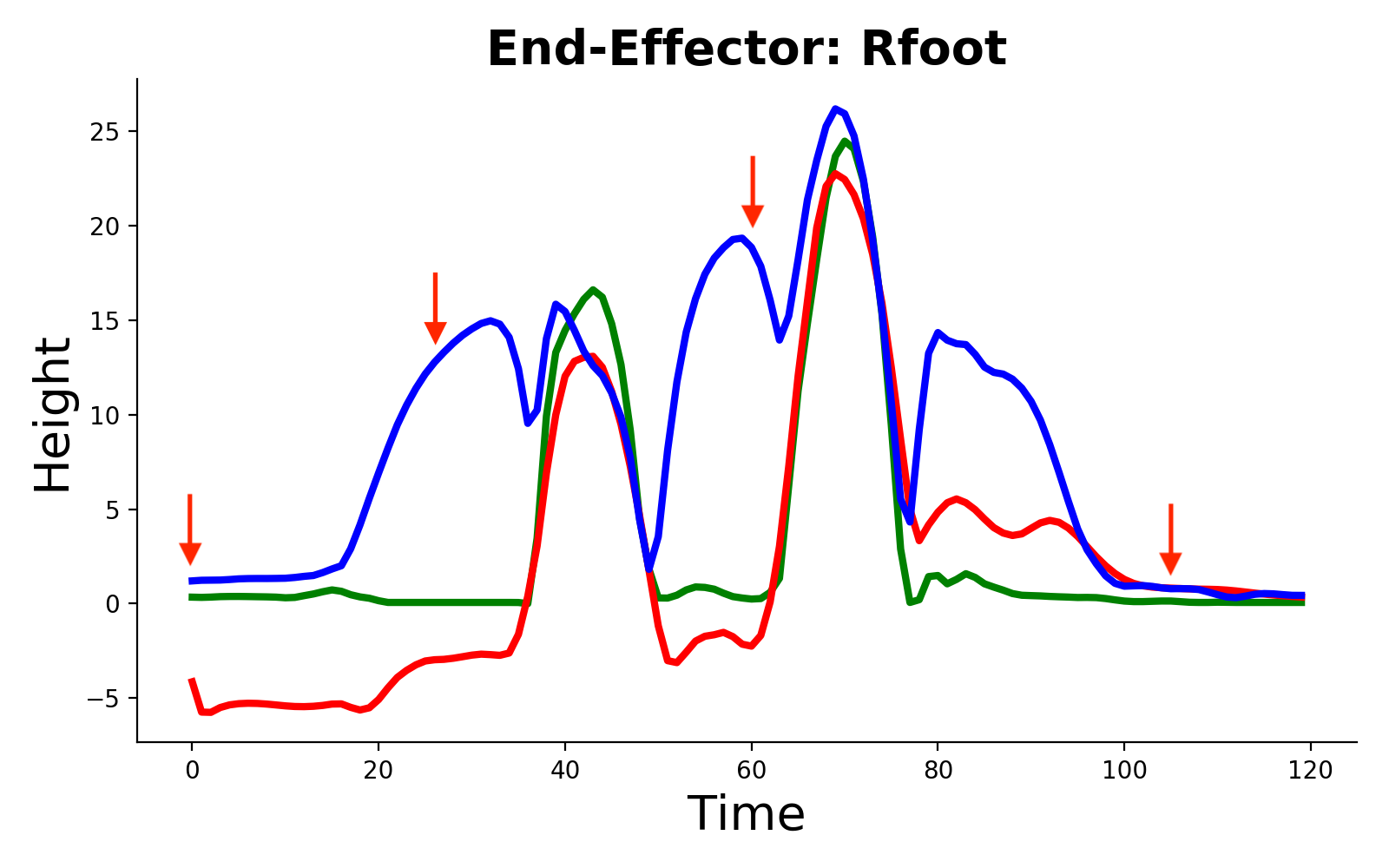} \hspace{5pt}
	    \includegraphics[width=.25\linewidth]{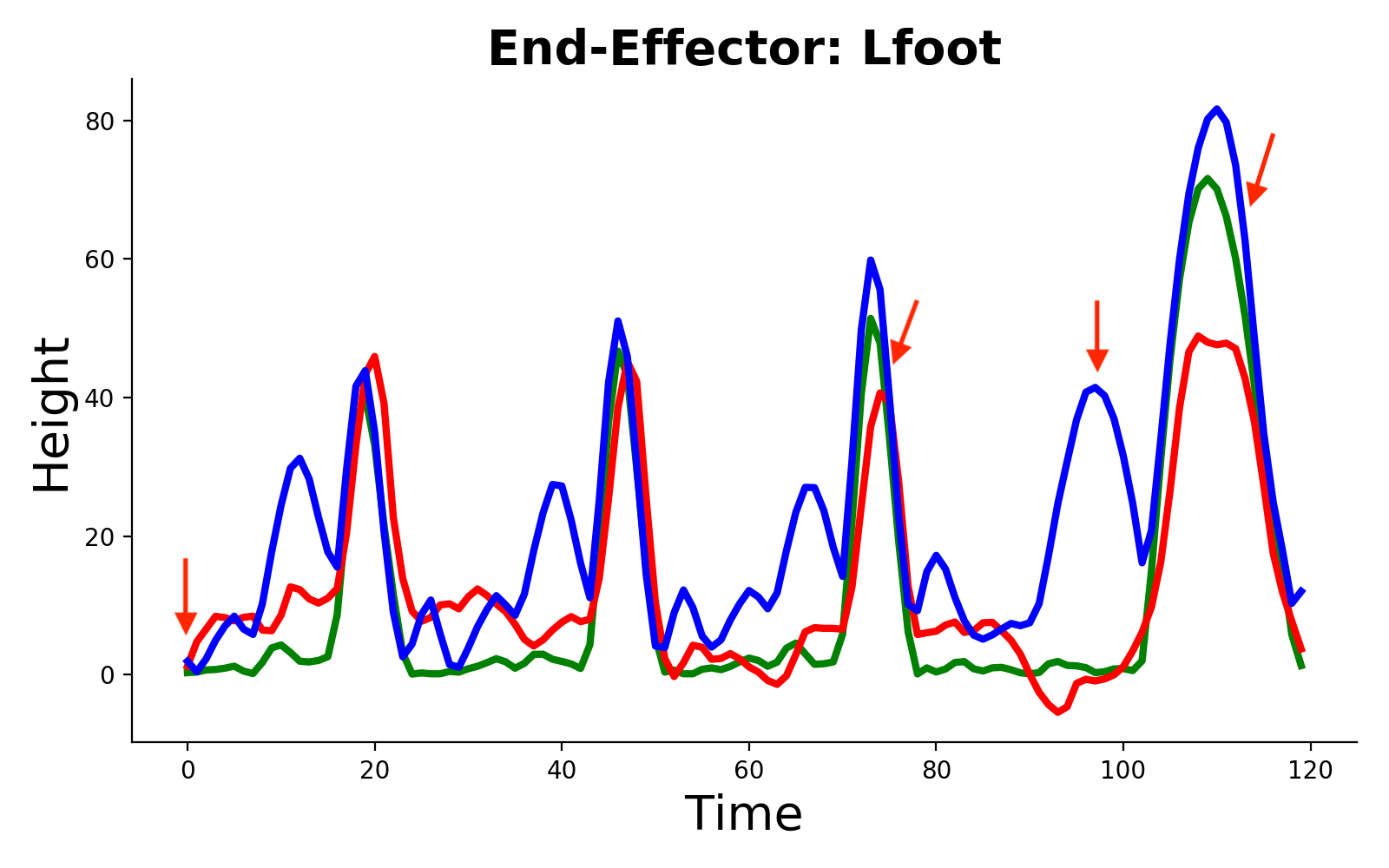} \hspace{-9pt}
	    \includegraphics[width=.25\linewidth]{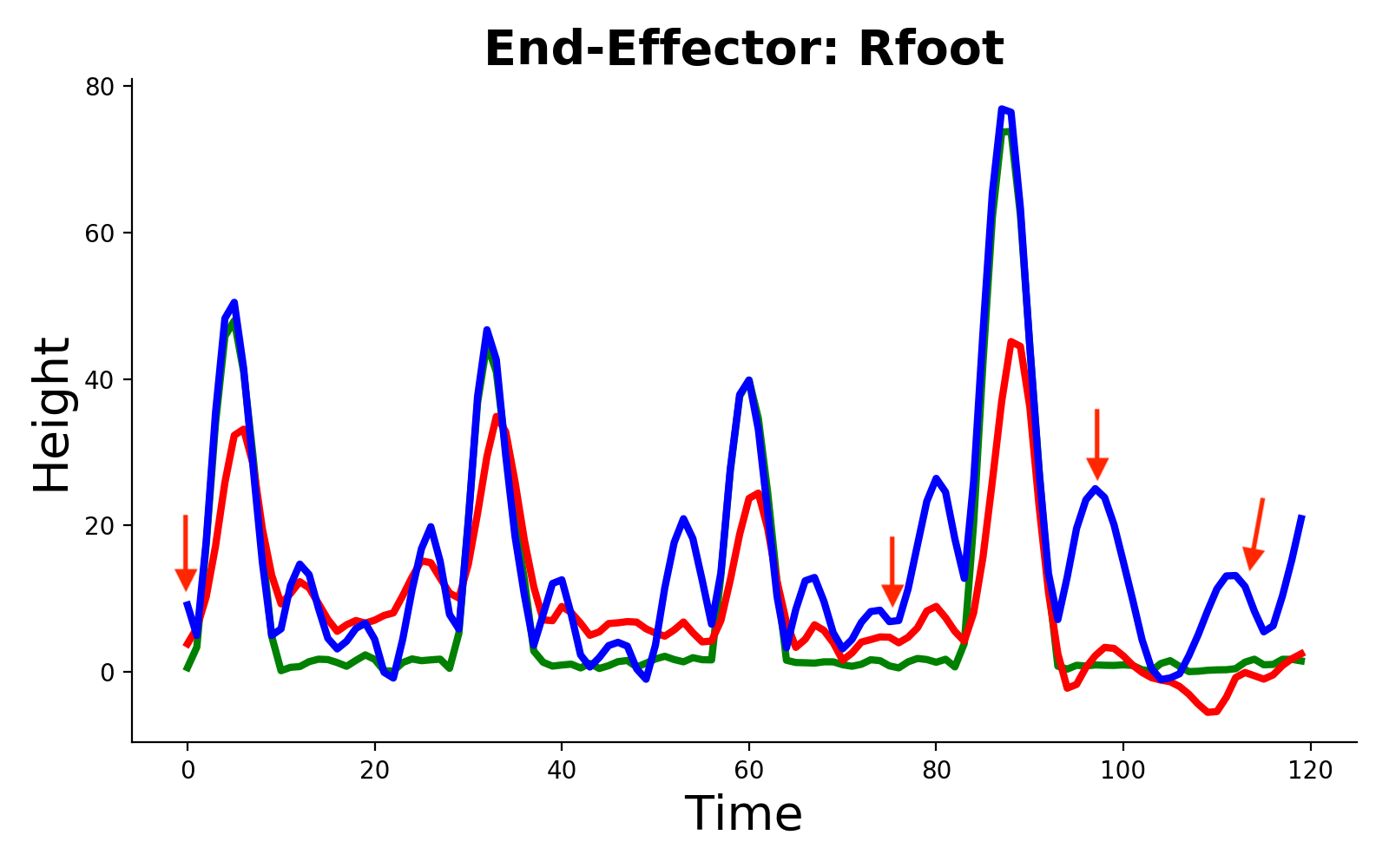} \hspace{-9pt} \\
	    \includegraphics[width=.25\linewidth]{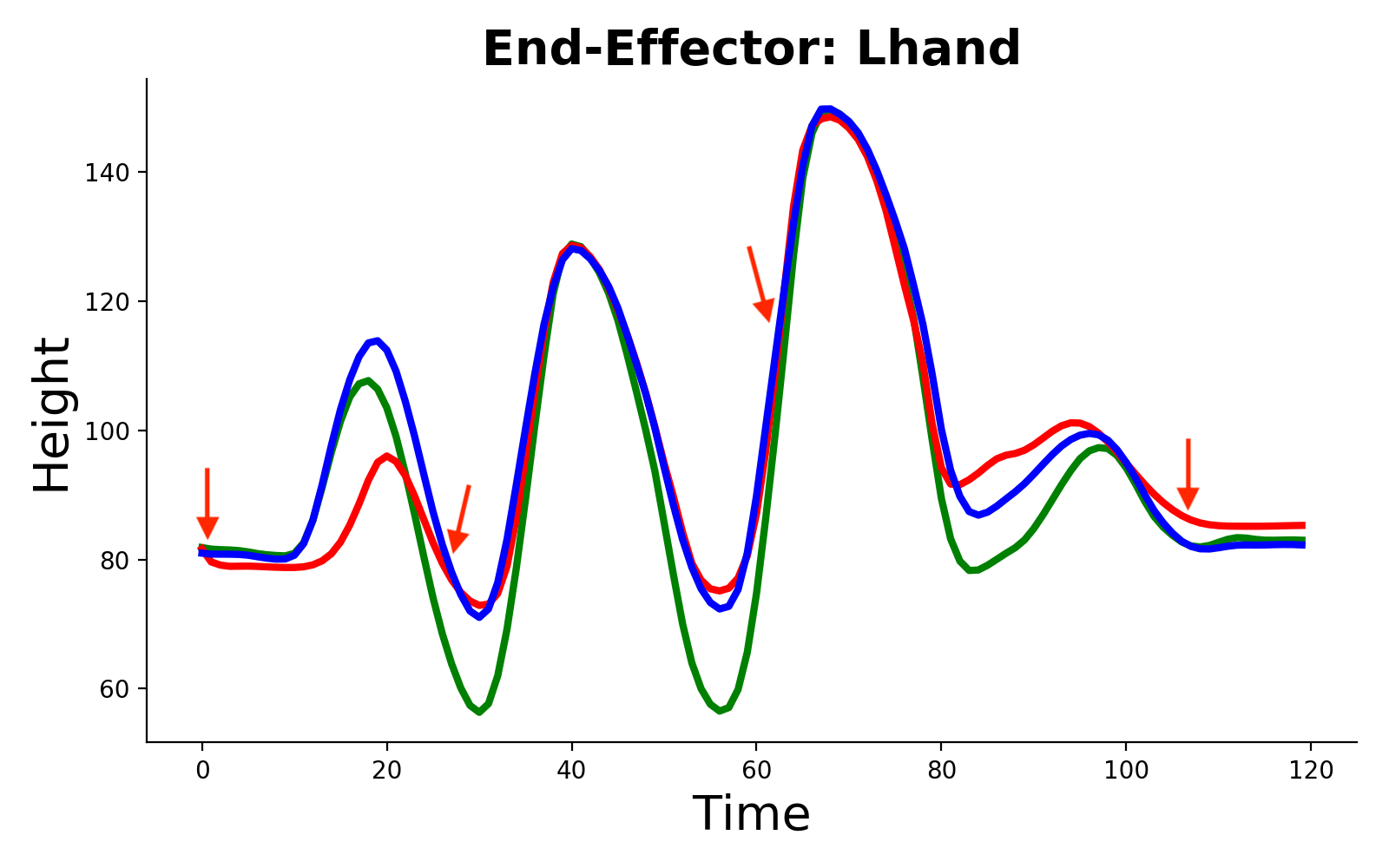} \hspace{-9pt}
	    \includegraphics[width=.25\linewidth]{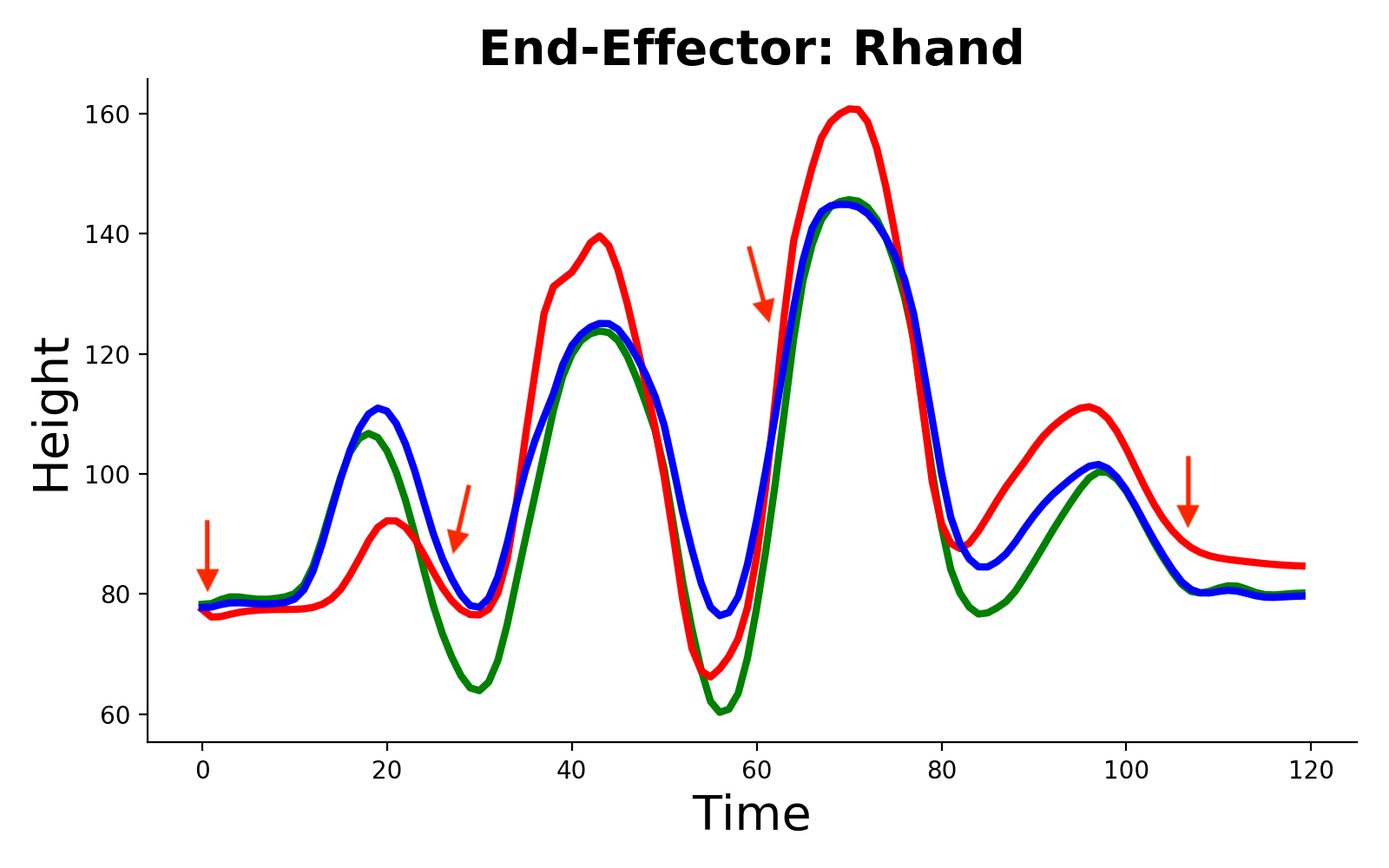} \hspace{5pt}
	    \includegraphics[width=.25\linewidth]{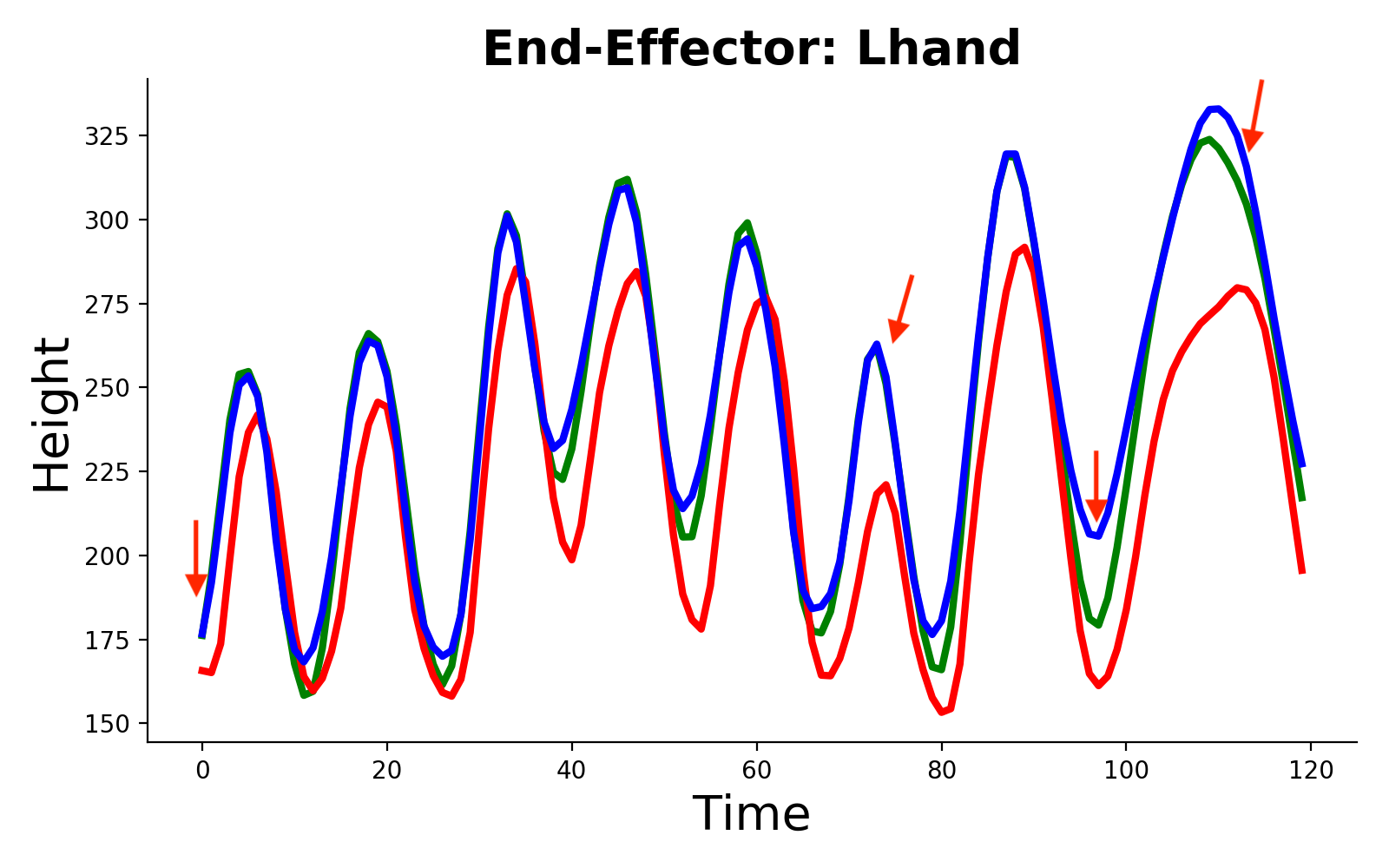} \hspace{-9pt}
	    \includegraphics[width=.25\linewidth]{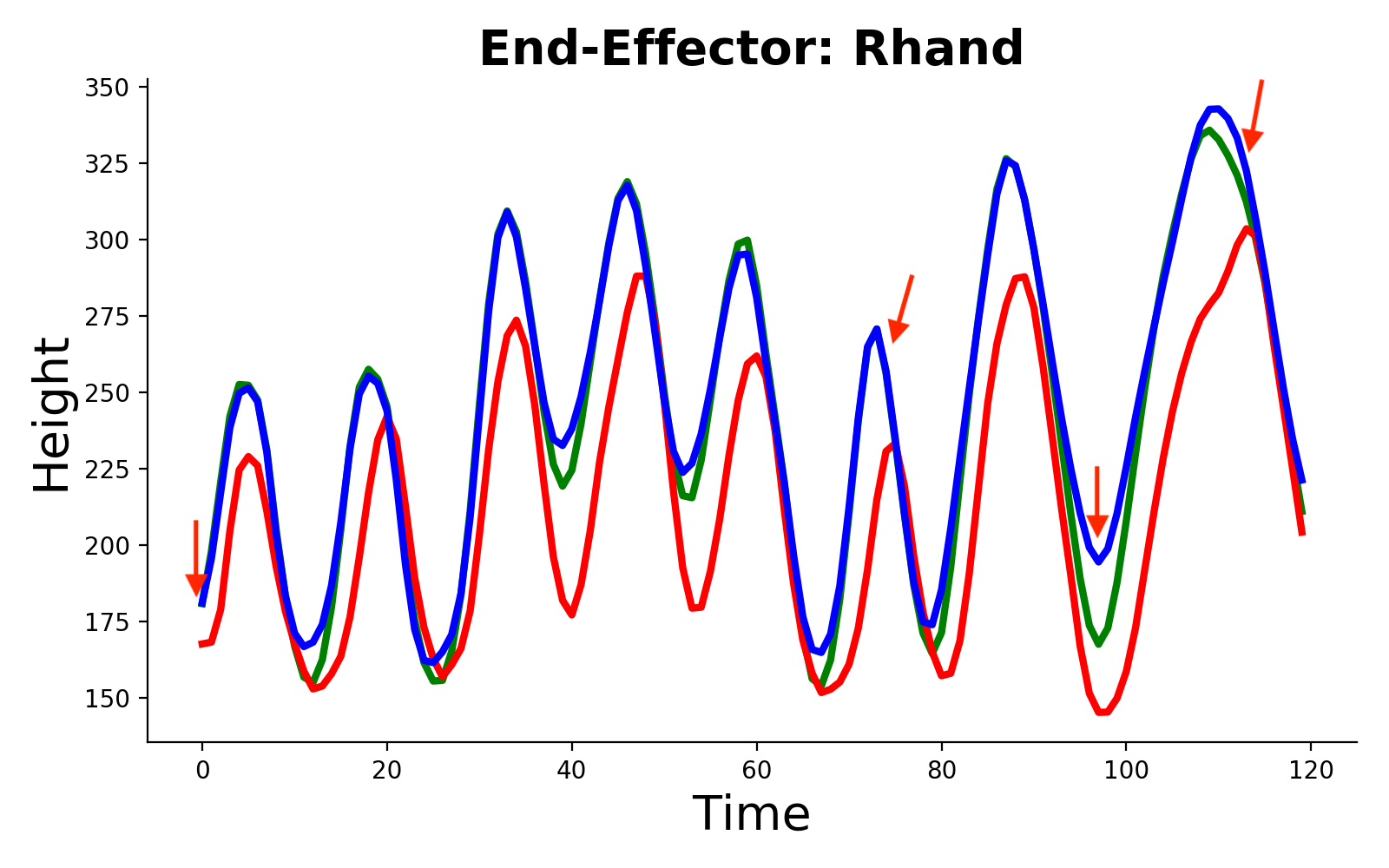} \hspace{-9pt}
	\end{subfigure}
	\end{center}
	\vspace{-.23in}
	\caption{Qualitative evaluation. We present a motion retargetting example of our method against the best baseline. Motion is retargetted from character Claire into Warrok Kurniawan (left) and Sporty Granny to Malcolm (right). Plots illustrating the left/right feet and hand end-effectors' height comparing against the groundtruth are shown at the bottom. Arrows in the plots determine the time steps of the shown animation frames. Please visit \url{goo.gl/mDTvem} for animated videos.}
	\vspace{-.1in}
\label{fig:qual_mixamo}
\end{figure*}

In this section we evaluate our method on the task of online motion retargetting, i.e., retargetting motion from one character to a target character as new motion frames are received. 
We present an ablation study to demonstrate the benefits of the different components of our method, and also compare against the previously described baselines.
In Table~\ref{table:online_mse}, we report the average MSE of the retargetted motion when our network is trained with different objectives: 
1) Our skeleton conditioned motion synthesis network (Section~\ref{sec:mo_synth}) trained with the autoencoder objective (i.e., input reconstruction) and the bone twisting constraint only.
2) Our network trained with the cycle consistency objective without adversarial training.
Specifically, the "otherwise" branch in Equation~\ref{eq:adv}, returns 0.
3) Our network trained with our full adversarial cycle consistency objective function which requires examples of motions performed by skeleton $B$ but not paired with any motions used as inputs during training.

\begin{table}[t] 
\centering
\small
\setlength{\tabcolsep}{3pt}
\begin{tabular}{l|c}
\Xhline{4\arrayrulewidth}
Method & MSE  \\
\Xhline{4\arrayrulewidth}
Ours: Autoencoder Objective & 10.25 \\
Ours: Cycle Consistency Objective & 8.51 \\
Ours: Adversarial Cycle Consistency Objective & \textbf{7.10}\\
\Xhline{4\arrayrulewidth}
Baseline: Conditional RNN & 13.65 \\
Baseline: Conditional RNN + Adv. Cycle Consistency & 26.93 \\
Baseline: Conditional MLP & 17.02 \\
Baseline: Conditional MLP + Adv. Cycle Consistency &  16.96\\
Baseline: Copy input quaternions and velocities & 9.00 \\

\Xhline{4\arrayrulewidth}
\end{tabular}
\vspace{-2pt}
\caption{Quantitative evaluation of online motion retargetting using mean square error (MSE).}
\vspace{-12pt}
\label{table:online_mse}
\end{table}

As it can be seen in Table~\ref{table:online_mse}, simply using the proposed FK layer within RNNs and training with an autoencoder objective (Ours: Autoencoder Objective), outperforms all standard neural network based baselines.
One explanation is that it is highly probable for the baselines to ignore the bone lengths of the target skeleton, and learn a motion representation that is dependent on the input skeleton.
The inability to disentangle motion properties from the input skeleton is more evident after training with our adversarial cycle consistency objective which still results in poor performance.
The inputs to the discriminator network are velocities, that is, local motion difference between adjacent time steps and global motion.
While this input contains information about the shift in joint locations through time, it does not capture any information about the spatial structure.
As a result, optimizing the baselines to fool the discriminator network, does not impose bone length constraints. Furthermore, encouraging velocities to be similar to the real data causes further bone length degradation (i.e., excessive stretching or shrinking) in absence of such constraints.
On the other hand, our architecture is designed to learn a skeleton invariant motion representation that can be directly transferred to the target skeleton through the FK layer.

\begin{figure*}[htp!]
\vspace*{-0.03in}
    \begin{center}
    \begin{subfigure}{1.0\linewidth}
	    \includegraphics[width=1.0\linewidth]{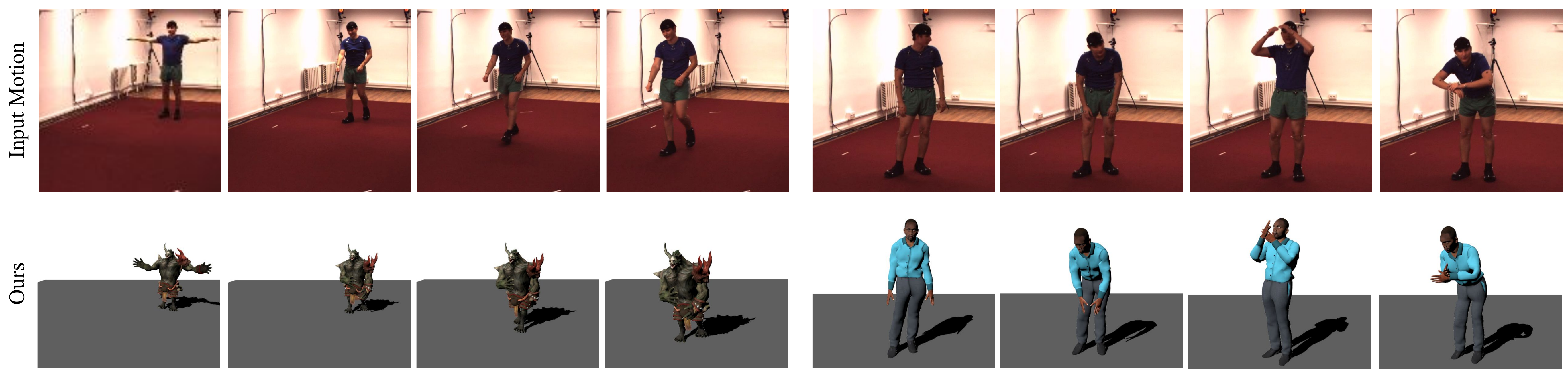}
	    \vspace{-9pt}
	\end{subfigure}
	\end{center}
	\vspace{-.23in}
	\caption{Qualitative evaluation on human videos. Motion is retargetted from estimated 3D pose from the Human 3.6M dataset into Mixamo 3D characters using the estimated 3D pose from \cite{martinez2017}. Please visit \url{goo.gl/mDTvem} for animated videos.}
	\vspace{-.1in}
\label{fig:qual_h36m}
\end{figure*}

The performance of our method improves when training our motion synthesis network with the proposed objectives for cycle consistency and adversarial cycle consistency.
While training with the autoencoder objective results in reasonable performance, often the network tries to match end-effector locations but does not fully capture the properties of the input motion. For example, when an input motion of a small character raising its hands is retargetted to a very tall character, the tall character is likely not able to raise its hands but only point in the same direction as the input motion.
Our network improves when trained with the cycle consistency objective alone. In the example of motion retargetting from a small to a tall character, cycle consistency loss prevents the tall character from directly matching end-effector positions of the small character as retargetting back to the small character would have resulted in stretching the limbs in the small character. The cycle consistency encourages the network to better learn the high level features of the input motion.

Finally, our method performs the best when our objective imposes both cycle consistency and realism via the full adversarial cycle consistency objective. The adversarial training helps the network to produce motions that cannot be distinguished from realistic motions of the target character. 

The baseline "Copy input quaternions and velocities" works better than the neural network baselines due to the following reasons:
1) Copying per-joint rotations of the input and performing forward kinematics already respects the target skeleton bone lengths, and 2) copying the velocities (i.e., global motion) avoids drifting that prediction models may suffer from. 
However, when retargetting motions between characters with significant skeleton difference, this baseline is prone to artifacts such as foot floating (see Figure~\ref{fig:qual_mixamo}).
This baseline is also not scalable to cases where different skeleton limits or topological structures are considered.

In Figure~\ref{fig:qual_mixamo}, we show qualitative results where we animate target characters using the output of our network using Blender~\cite{blender}, a character animation software.
For all the joints that are not modeled by our network (e.g., the fingers), we simply directly copy the joint rotations from the input motion if the corresponding joint names match in the input and the target skeleton, otherwise we leave them fixed. 

\vspace*{-0.03in}
\subsection{Online Motion Retargetting from Human Video}
\vspace*{-0.03in}

In this section we present motion retargetting from human video input into characters using the model trained from the Mixamo data only.
We use the Human 3.6M videos as input, the algorithm from \cite{martinez2017} to estimate the 3D pose of each frame, and the ground truth 3D skeleton root displacement (3D pose estimation algorithms usually assume the person is centered).
The videos are subsampled to 25 FPS, and the estimated 3D poses are processed similar to our previous experiment.
The algorithm in \cite{martinez2017} only outputs 17 joints compared to the 22 joints needed by our network.
Therefore, we manually map the 17 joints to 22 by duplicating the following joint positions in Human 3.6M to corresponding Mixamo joints:
Spine into Spine and Spine1,  LeftShoulder into LeftShoulder and LeftArm, RightShoulder into RightShoulder and RightArm , LeftFoot into LeftFoot and LeftToeBase, RightFoot into RightFoot and RightToeBase.
Note that this mapping will create bones of zero length during test time.
Thus, our network essentially only sees 17 joints but uses 22 joints as input.
During visualization, we do not rotate joints that are not predicted by our network (i.e., fingers). 
As shown in Figure~\ref{fig:qual_h36m}, our network is able to generalize to never-seen skeletons and motions estimated from monocular human videos.
More video results and analyses are included in supplementary materials.

%% file: 6_conclusion.tex
\vspace*{-0.03in}
\section{Conclusion and Future Work}
\label{sec:conclusion}
\vspace*{-0.04in}

We have presented a neural kinematic network with an adversarial cycle consistency training objective for motion retargetting.
Our network only observes a sequence of \text{xyz}-coordinates of joints from existing animations, motion capture or 3D pose estimates of monocular human videos, and transfers the motion to a target humanoid character without risking skeleton deformations that occur in the baselines.
The success of our method attributes to the following factors:
1) The proposed $\textit{Forward Kinematics layer}$ helps to discover joint rotations of target skeleton that are independent of the input skeleton.
2) The cycle consistency of the retargetting objective prevents regressing to the end-effector positions of the input motion.
3) The adversarial objective helps the network to produce realistic motions.
4) The bone twist loss constrains the solution space of $\textit{Inverse Kinematics}$ and prevents bone twisting in the retargetted motion.

Our current method has limitations.
First, we perform retargetting on a fixed number of joints.
Handling a variable number of joints is challenging as the retargetting algorithm is expected to automatically select end-effectors of interest when transferring motions.
Second, we assume the environment in which the target character is being animated lacks physical constraints such as gravity.
Future work will include equipping the network with physics simulators to generate more natural and physically plausible movements of the target characters with different muscle/bone mass distributions.
Third, the input to our method still requires 3D information (\text{xyz}-coordinates of joints). 
Future work will also include training our network end-to-end by using monocular videos as input. That may require the algorithm to learn view-invariant features.

\vspace*{-0.1in}
\paragraph{Acknowledgments.} 
This work was supported in part by Gift from Adobe Research, NSF CAREER IIS-1453651, and Rackham Merit Fellowship. We would like to thank Aaron Hertzmann for helpful discussions.

%% file: 7_supp.tex
\clearpage
\newpage
\onecolumn
\section*{\fontsize{15}{18}\selectfont Appendix}
\begin{appendix}

\section{Quantitative Evaluation per Motion Retargetting Scenario, and Analysis}
\noindent In this section, we present quantitative evaluation for the different motion retargetting scenarios mentioned in the main text.
We then present findings showing how our method significantly outperforms the best performing baseline (copy quaternions and velocities).

\vspace{10pt}
\noindent In Table~\ref{supp:known_motion}, we show results of retargetting motion previously seen during training into two target scenarios: 1) Character has been seen during training, but not performing the motion used as input (left). 2) Character has never been seen during training (right).
In Table~\ref{supp:new_motion}, we show results of retargetting motion never seen during training into two target scenarios: 1) Character that has been seen during training (left). 2) Character has never been seen during training (right).

\vspace{15pt}
\begin{table}[ht]
\centering
\begin{tabular}{lr}

\setlength{\tabcolsep}{3pt}
\begin{tabular}{c}
\Xhline{4\arrayrulewidth}
Known Motion / Known Character \\
\begin{tabular}{l|c}
\Xhline{4\arrayrulewidth}
Method & MSE  \\
\Xhline{4\arrayrulewidth}
Ours: Autoencoder Objective & 8.61  \\
Ours: Cycle Consistency Objective & 5.68  \\
Ours: Adversarial Cycle Consistency Objective & 5.35 \\
\Xhline{4\arrayrulewidth}
\end{tabular} \\
Ground-truth joint location variance through time: 4.8 \\
\Xhline{4\arrayrulewidth}
\end{tabular}
\vspace{-2pt}
&

\setlength{\tabcolsep}{3pt}
\begin{tabular}{c}
\Xhline{4\arrayrulewidth}
Known Motion / New Character \\
\begin{tabular}{l|c}
\Xhline{4\arrayrulewidth}
Method & MSE  \\
\Xhline{4\arrayrulewidth}
Ours: Autoencoder Objective & 2.16  \\
Ours: Cycle Consistency Objective & 1.55  \\
Ours: Adversarial Cycle Consistency Objective & 1.35 \\
\Xhline{4\arrayrulewidth}
\end{tabular} \\
Ground-truth joint location variance through time: 1.5 \\
\Xhline{4\arrayrulewidth}
\end{tabular}
\vspace{-2pt}

\end{tabular}
\caption{Quantitative evaluation of online motion retargetting using mean square error (MSE). Case study: Known motion / known character (left), and known motion / new character (right).}
\label{supp:known_motion}
\end{table}



\begin{table}[ht]
\centering
\begin{tabular}{lr}

\setlength{\tabcolsep}{3pt}
\begin{tabular}{c}
\Xhline{4\arrayrulewidth}
New Motion / Known Character \\
\begin{tabular}{l|c}
\Xhline{4\arrayrulewidth}
Method & MSE  \\
\Xhline{4\arrayrulewidth}
Ours: Autoencoder Objective & 6.55  \\
Ours: Cycle Consistency Objective & 4.38  \\
Ours: Adversarial Cycle Consistency Objective & 4.39 \\
\Xhline{4\arrayrulewidth}
\end{tabular} \\
Ground-truth joint location variance through time: 3.6 \\
\Xhline{4\arrayrulewidth}
\end{tabular}
\vspace{-2pt}

&

\setlength{\tabcolsep}{3pt}
\begin{tabular}{c}
\Xhline{4\arrayrulewidth}
New Motion / New Character \\
\begin{tabular}{l|c}
\Xhline{4\arrayrulewidth}
Method & MSE  \\
\Xhline{4\arrayrulewidth}
Ours: Autoencoder Objective & 24.16  \\
Ours: Cycle Consistency Objective & 23.49  \\
Ours: Adversarial Cycle Consistency Objective & 18.02 \\
\Xhline{4\arrayrulewidth}
\end{tabular} \\
Ground-truth joint location variance through time: 11.6 \\
\Xhline{4\arrayrulewidth}
\end{tabular}
\vspace{-2pt}
\end{tabular}
\caption{Quantitative evaluation of online motion retargetting using mean square error (MSE). Case study: New motion / known character (left), and new motion / new character (right).}
\label{supp:new_motion}
\end{table}

\vspace{15pt}

\noindent From these results, we can observe the benefits of our full adversarial cycle training vs only using cycle training.
In both input motion scenarios --- seen during training and never seen during training --- retargetting into a never before seen target skeleton results in overall performance improvement.
For known input motions, retargetting into a new character results in a performance improvement of $\mathbf{12.9\%}$, while retargetting into a known character results in a performance improvement of $\mathbf{5.8\%}$.
Additionally, for new input motions, retargetting into a new character results in a performance improvement of $\mathbf{23.3\%}$, while retargetting into a known character results in a similar performance.
In the previous analysis, we can clearly see that learning character behaviors in the training data results in an overall performance boost when the target character has been seen before.
Most importantly, learning skeleton conditioned behaviors results in much better generalization to new characters compared to training with cycle alone.
However, we can also see that some scenarios reflect larger errors than others.
To explain this phenomenon, we measure the movement in the ground-truth motion sequences by computing the average character height normalized joint location variance through time presented at the bottom of each table.
This result shows that the more movement there is in the ground-truth sequence, the larger the MSE becomes, thus the larger errors seen on some of the test scenarios previously presented.

\vspace{5pt}
\newpage

\noindent Next, we quantitative evaluate our method and the best performing baseline (copy input quaternions and velocities) by separating testing examples into bins based on average movement through time observed in the ground-truth target motion.
This evaluation gives us a clearer insight on how much input movement in space each method can handle during retargetting, and how our method is outperforming the best baseline.

\vspace{10pt}
\begin{figure}[h]
    \begin{center}
    \begin{subfigure}{0.6\linewidth}
	    \includegraphics[width=1.0\linewidth]{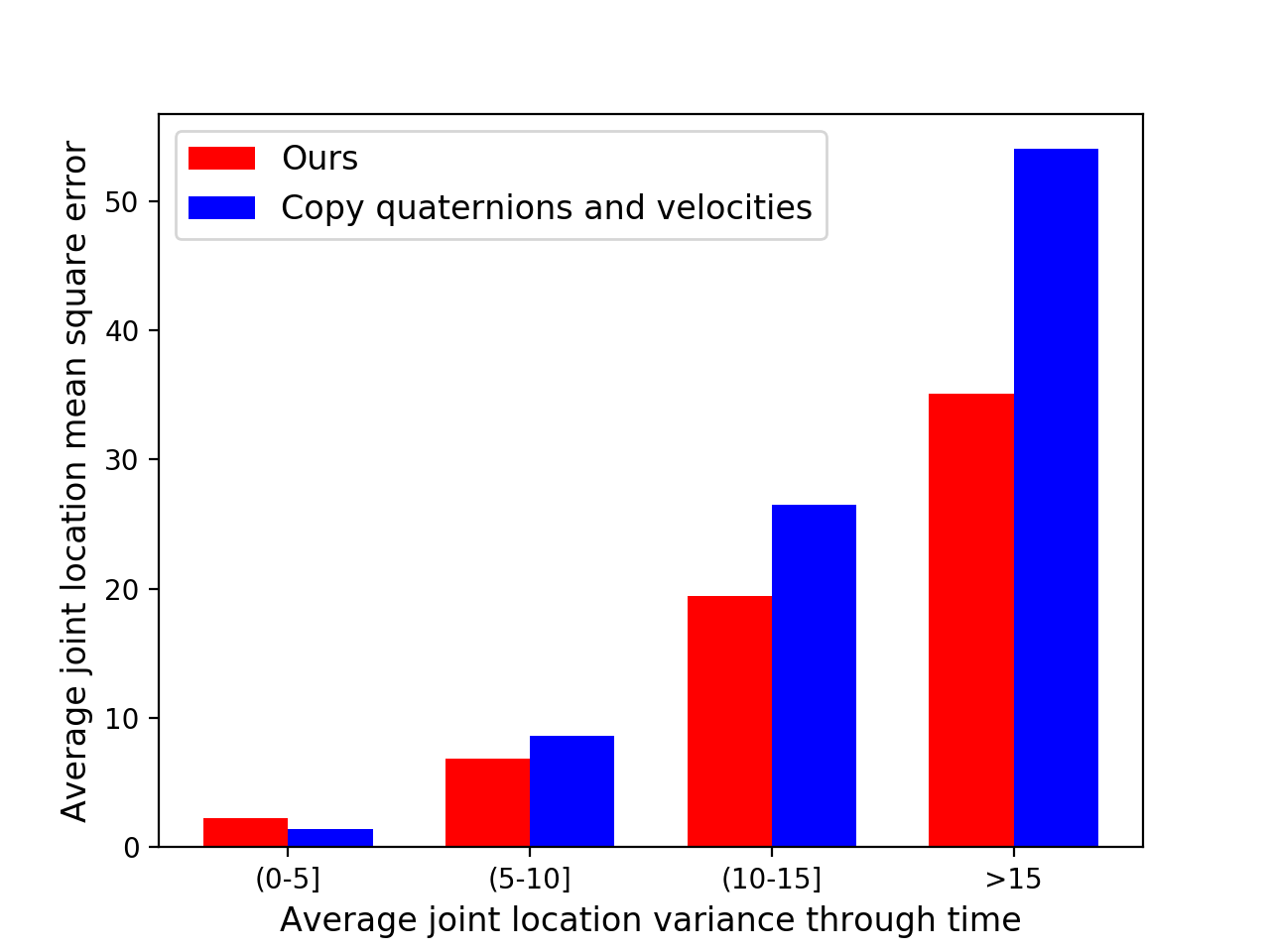}
	    \vspace{-9pt}
	\end{subfigure}
	\end{center}
	\vspace{-15pt}
	\caption{Quantitative evaluation based on movement through time. The vertical axis denotes mean square error, and the horizontal axis denotes the \textit{xyz}-coordinate average variance through time observed in the ground-truth. The average joint location variance is normalized by character height.}
\label{supp:vels_eval}
\end{figure}
\vspace{15pt}

\noindent In Figure~\ref{supp:vels_eval}, we can observe that our method outperforms the baseline as the movement in the evaluation videos increase.
Our method substantially outperforms the baseline when the average joint location variance is larger than 5, however, the baseline marginally outperforms our method when the average joint location variance is less than or equals to 5.
This result shows that by simply copying the input motion into the target character we cannot guarantee motion retargetting that follows the correct motion in the target character.
Therefore, we have to rely on a model that has understanding of the target character and input motion relationships for synthesizing skeleton conditioned motion.

\newpage

\section{Denoising 3D Pose Estimation by Motion Retargetting}
\noindent In this section, we show the denoising power of our model on estimated 3D poses.
Most 3D pose estimation algorithms do it in a per-frame manner completely ignoring temporal correlations among the estimated poses in videos at every time step.
We use our method trained on the Mixamo dataset to retarget the 3D pose estimated by [16] back into the input motion skeleton (Human 3.6M skeleton) to demonstrate denoising effects on the input pose sequence.
We compute the Human 3.6M skeleton from the first frame pose in the sequence $\texttt{WalkTogether 1.60457274}$ performed by $\texttt{Subject 9}$.
We evaluate for denoising by plotting the end-effector height trajectories (hands and feet) of the local motion output of our method since the algorithm we use to estimate 3D pose assumes centered human input.
Below we plot end-effector trajectories of our retargetted poses and the originally estimated poses (input to our method) for selected examples (None: Please check our project website for better appreciation of the denoising happening \url{goo.gl/mDTvem}).

\begin{figure}[h!]
    \begin{center}
	\begin{subfigure}{1.0\linewidth}
	    \includegraphics[width=.25\linewidth]{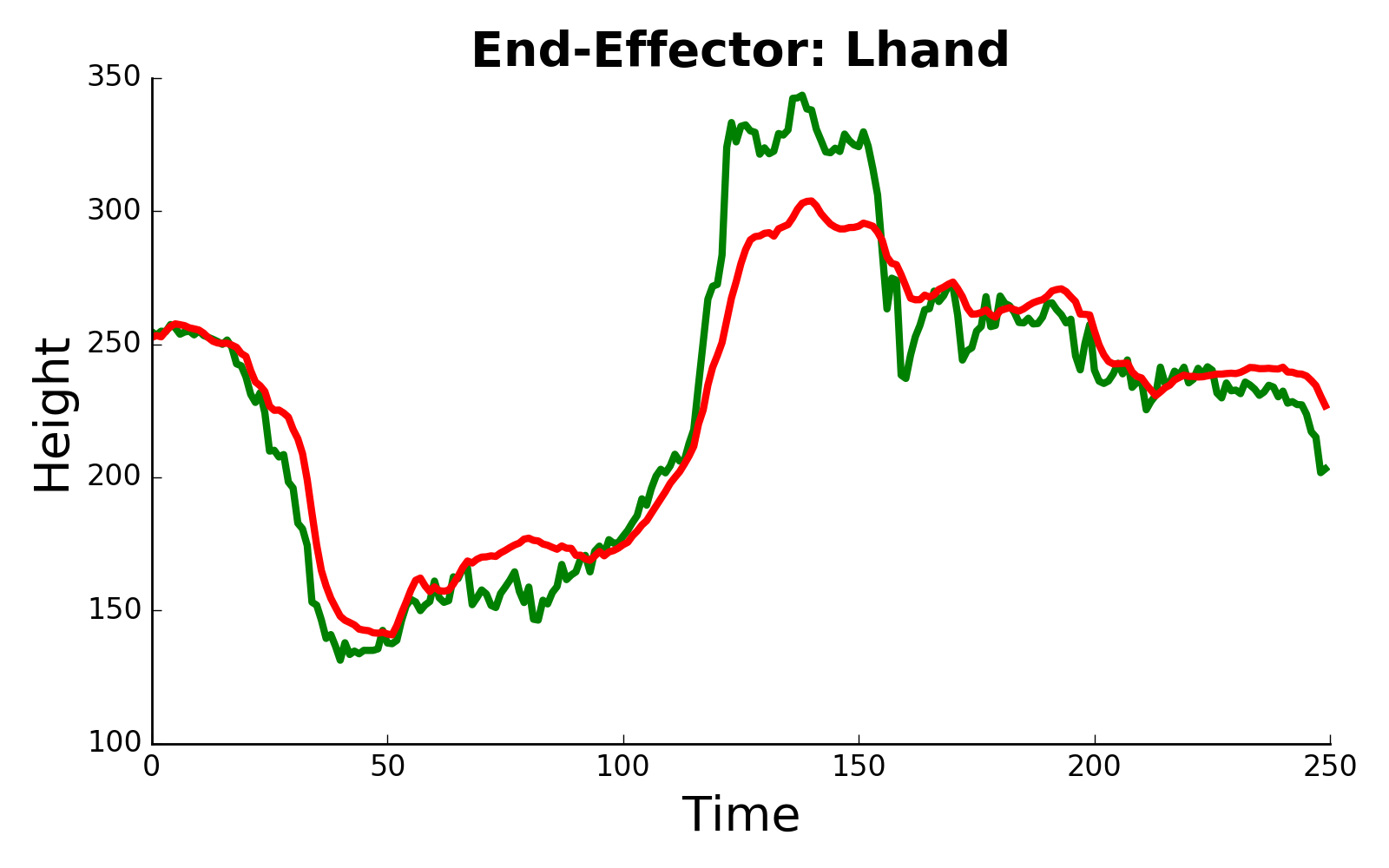} \hspace{-5pt}
	    \includegraphics[width=.25\linewidth]{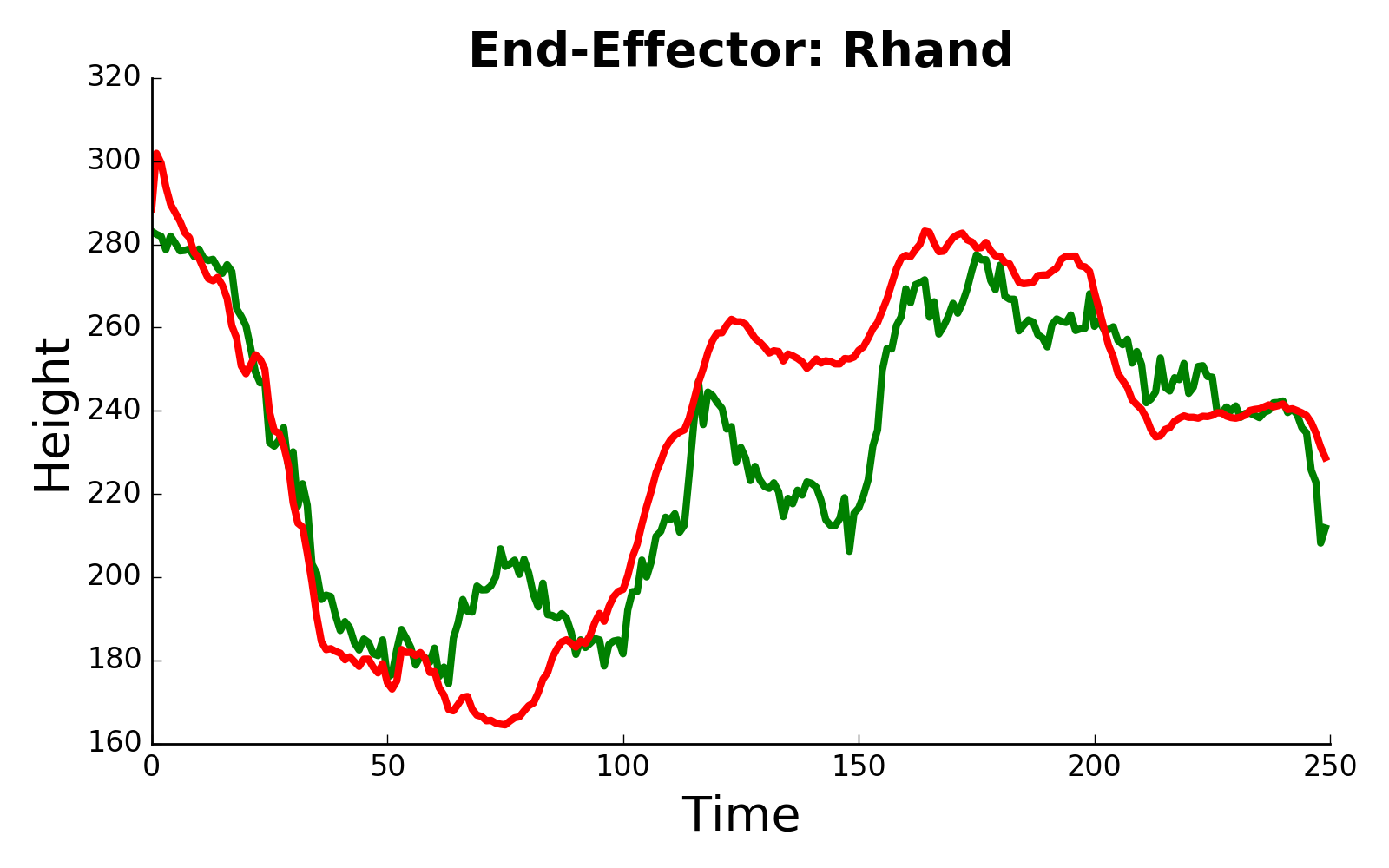} \hspace{-5pt}
	    \includegraphics[width=.25\linewidth]{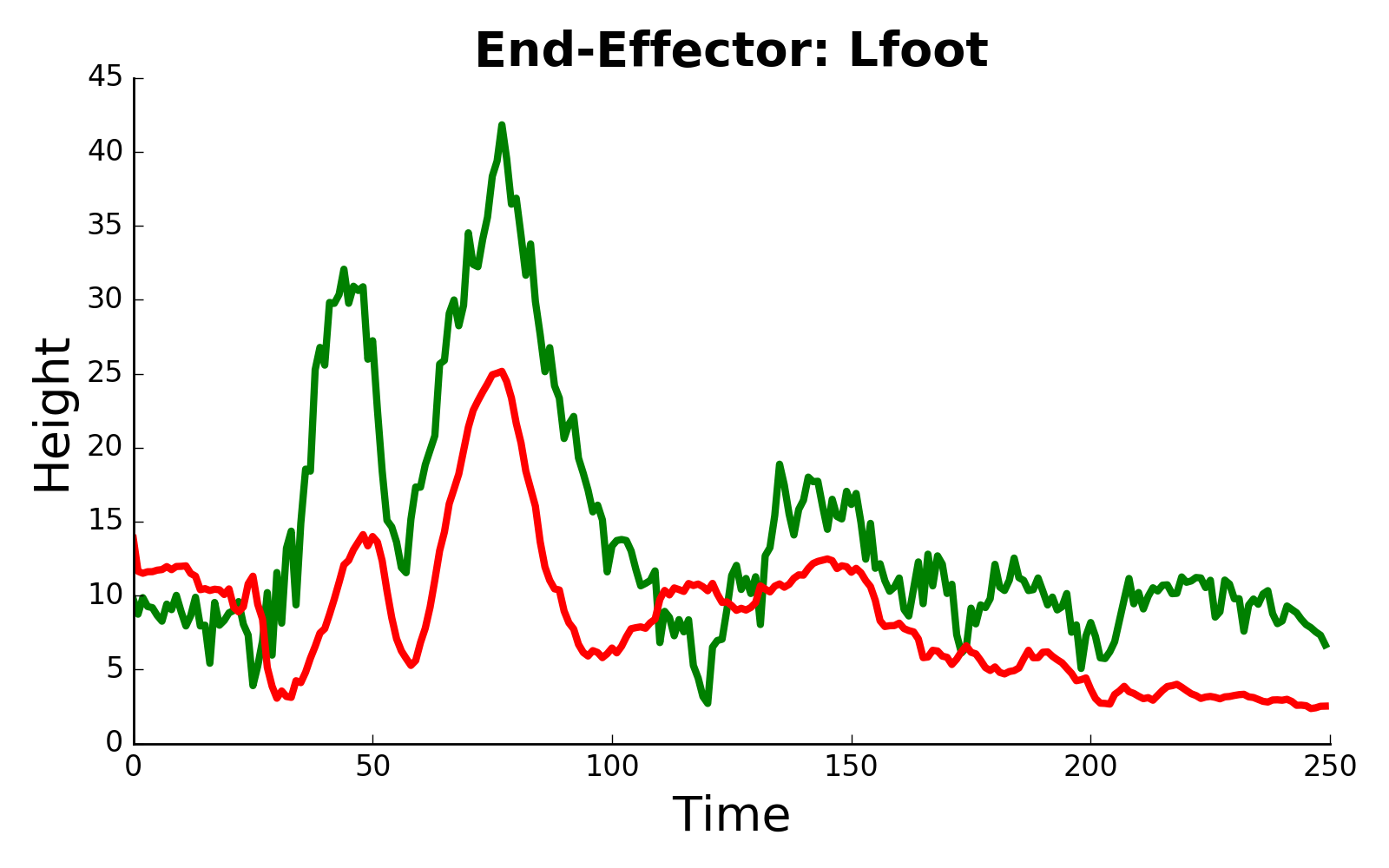} \hspace{-5pt}
	    \includegraphics[width=.25\linewidth]{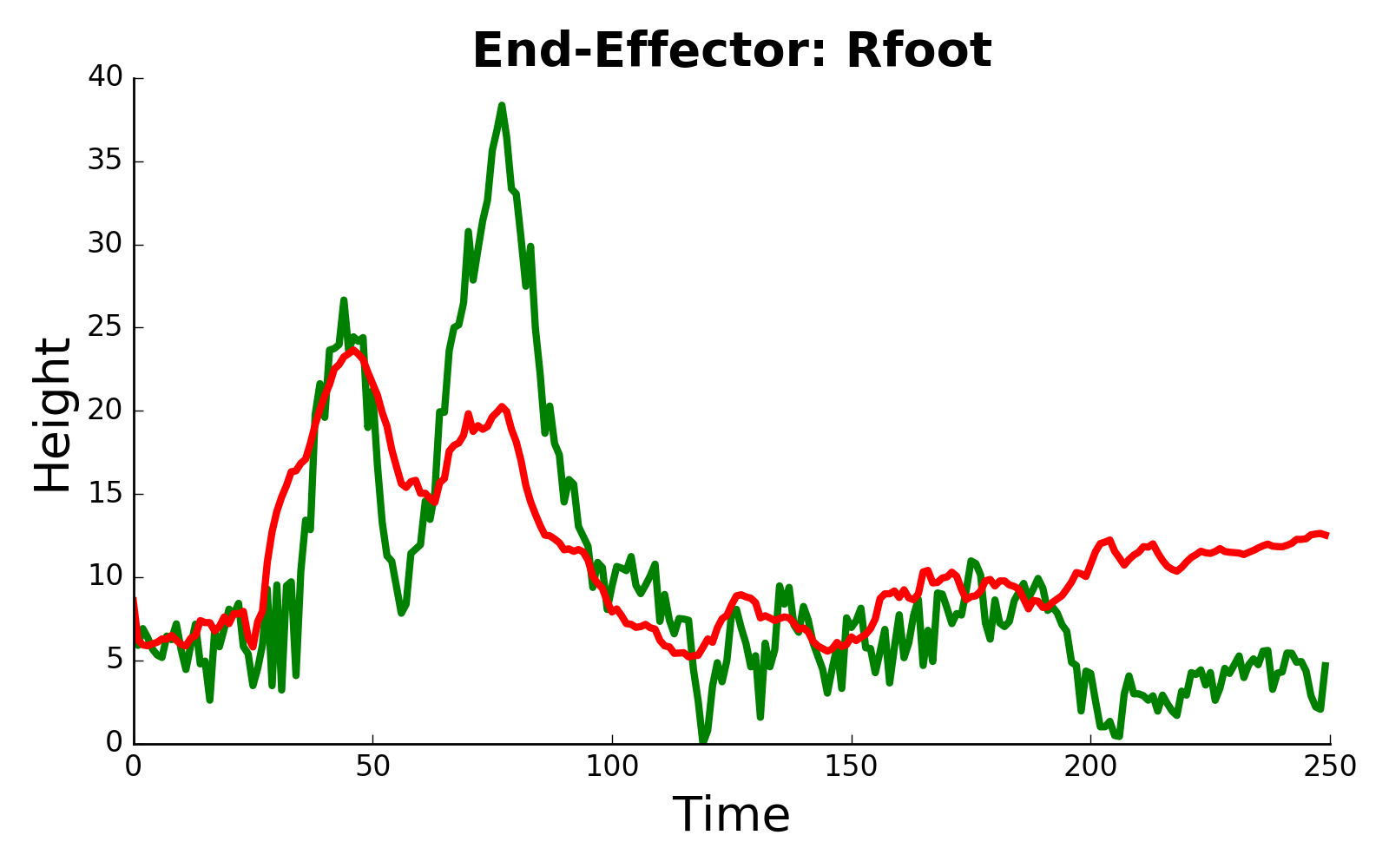} \hspace{-5pt} \\

	    \includegraphics[width=.25\linewidth]{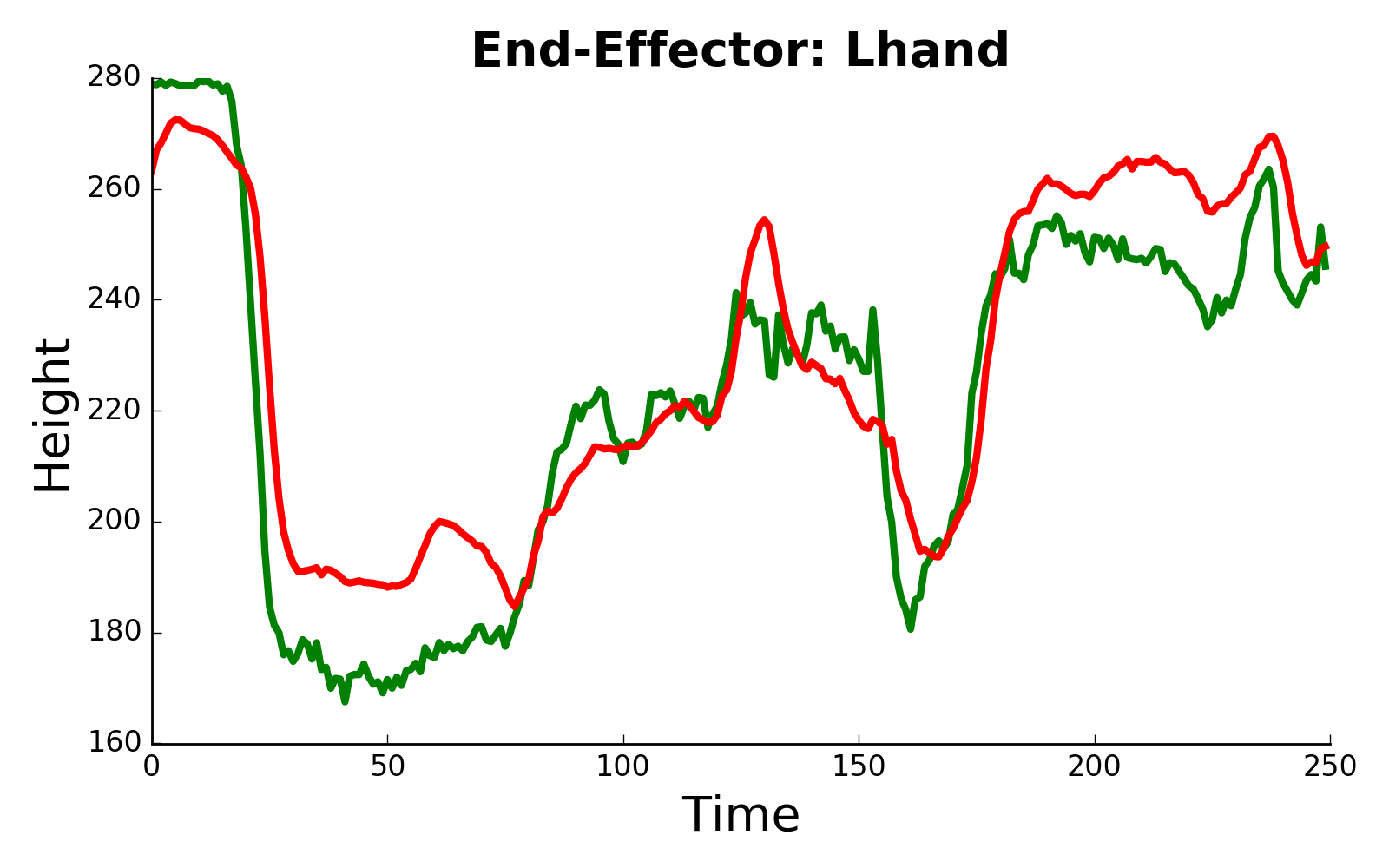} \hspace{-5pt}
	    \includegraphics[width=.25\linewidth]{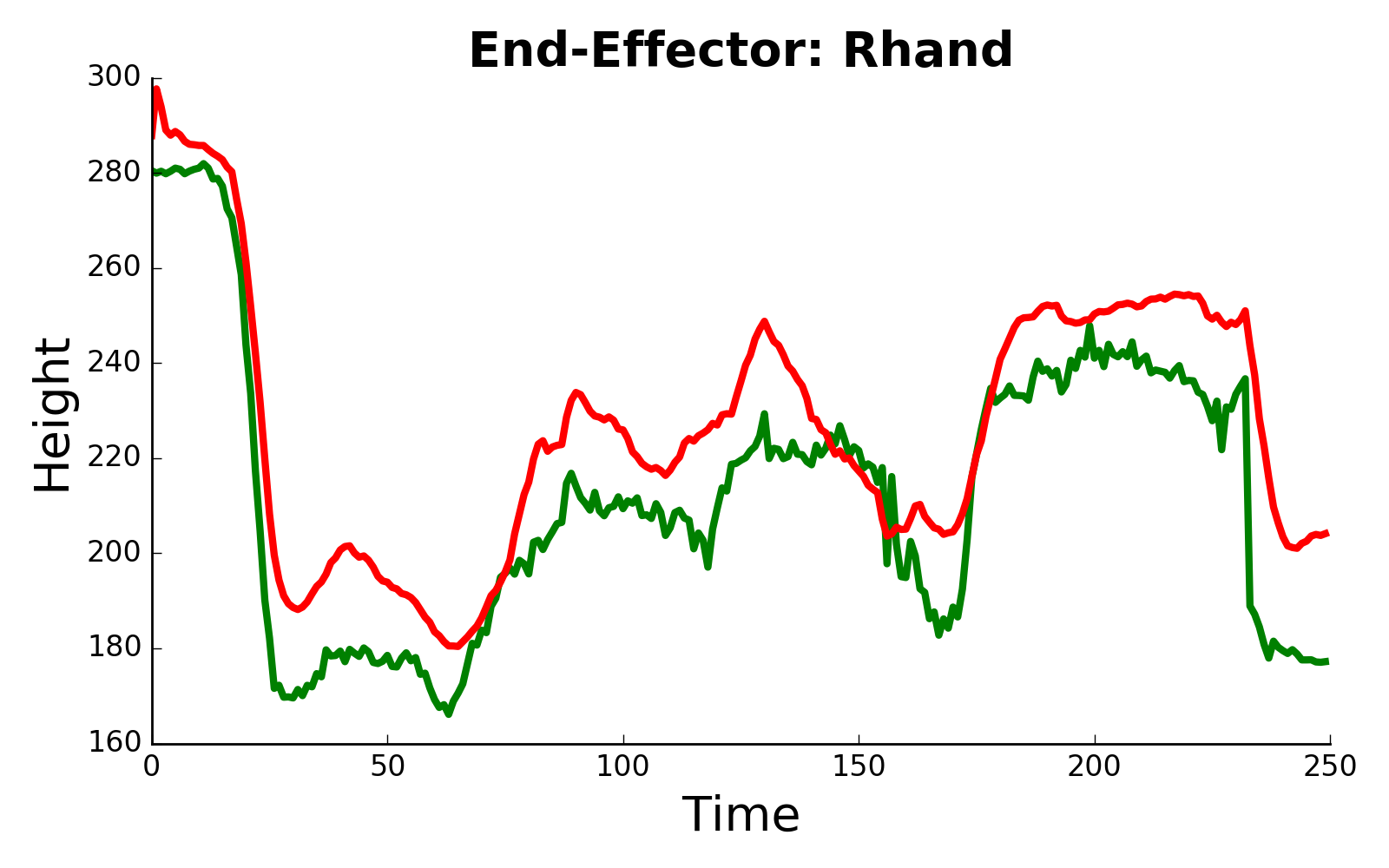} \hspace{-5pt}
	    \includegraphics[width=.25\linewidth]{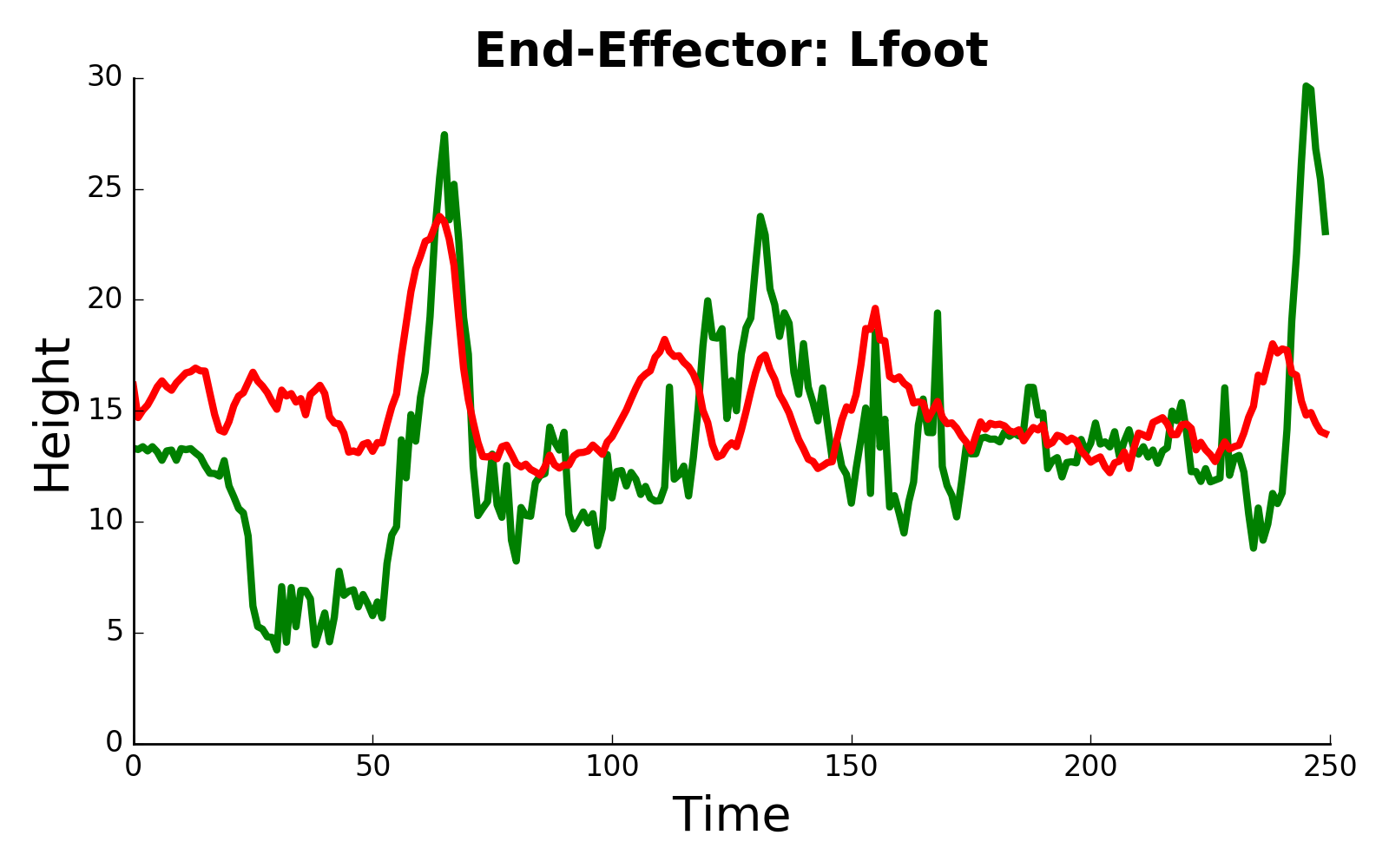} \hspace{-5pt}
	    \includegraphics[width=.25\linewidth]{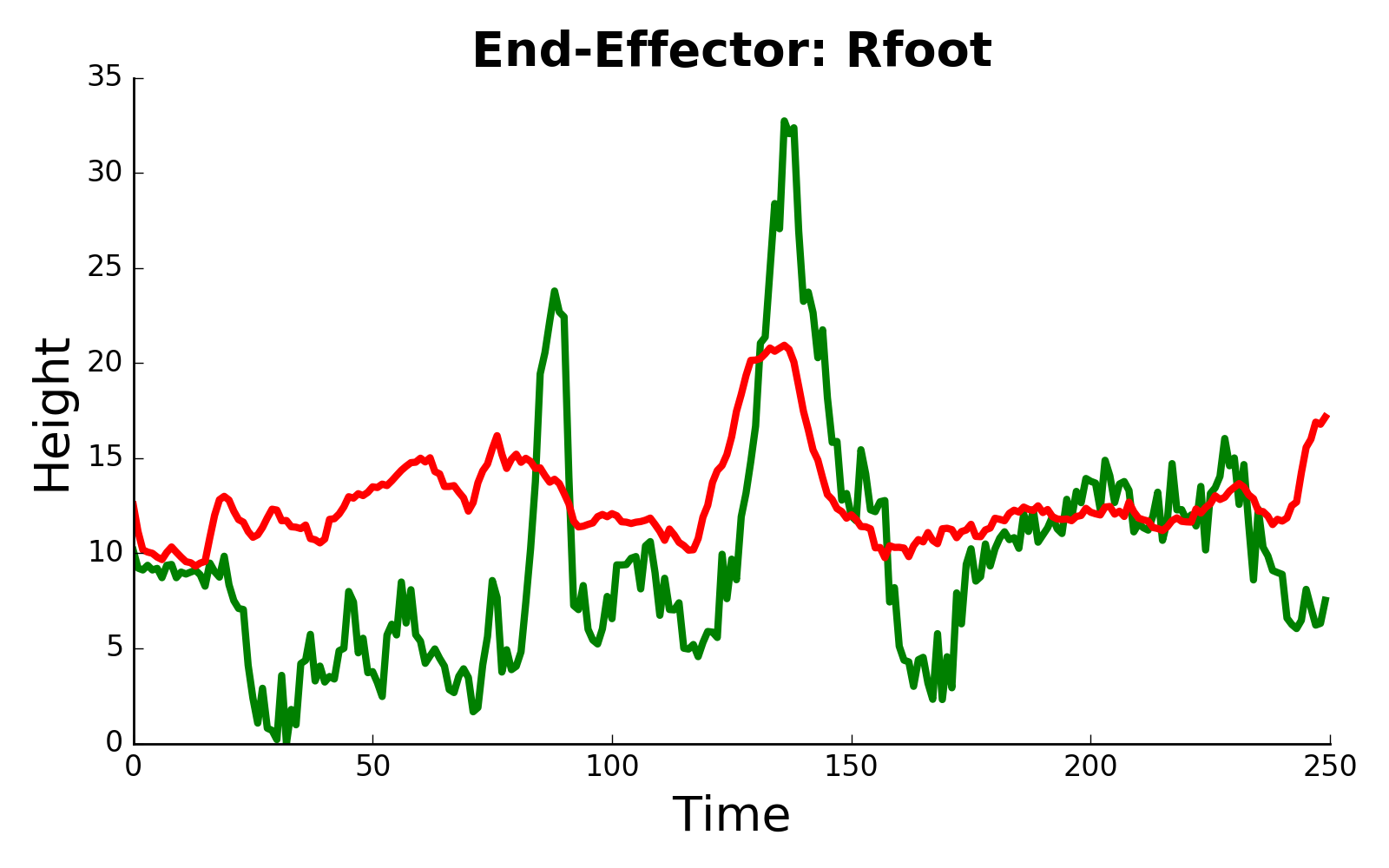} \hspace{-5pt} \\
	    
	    \includegraphics[width=.25\linewidth]{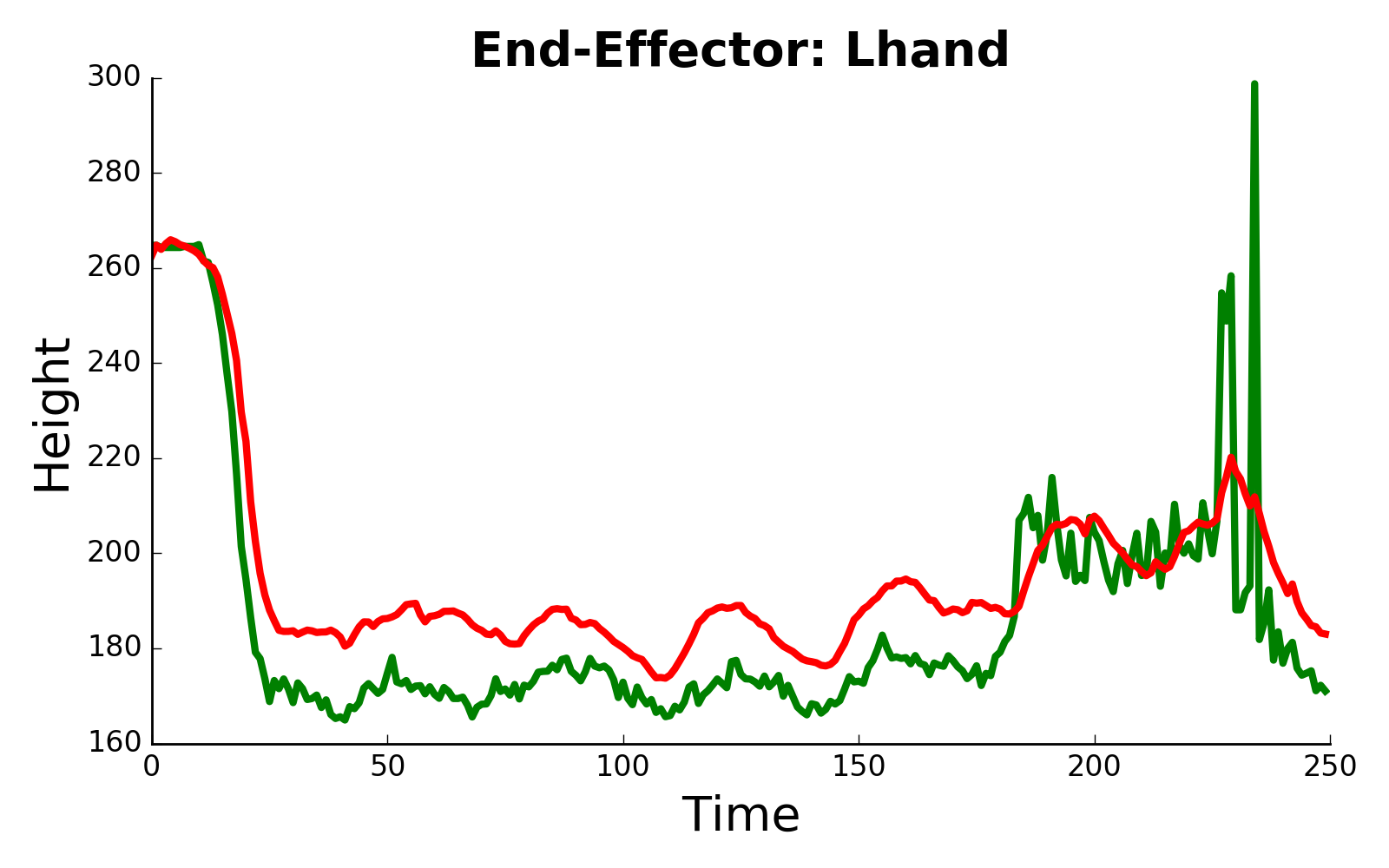} \hspace{-5pt}
	    \includegraphics[width=.25\linewidth]{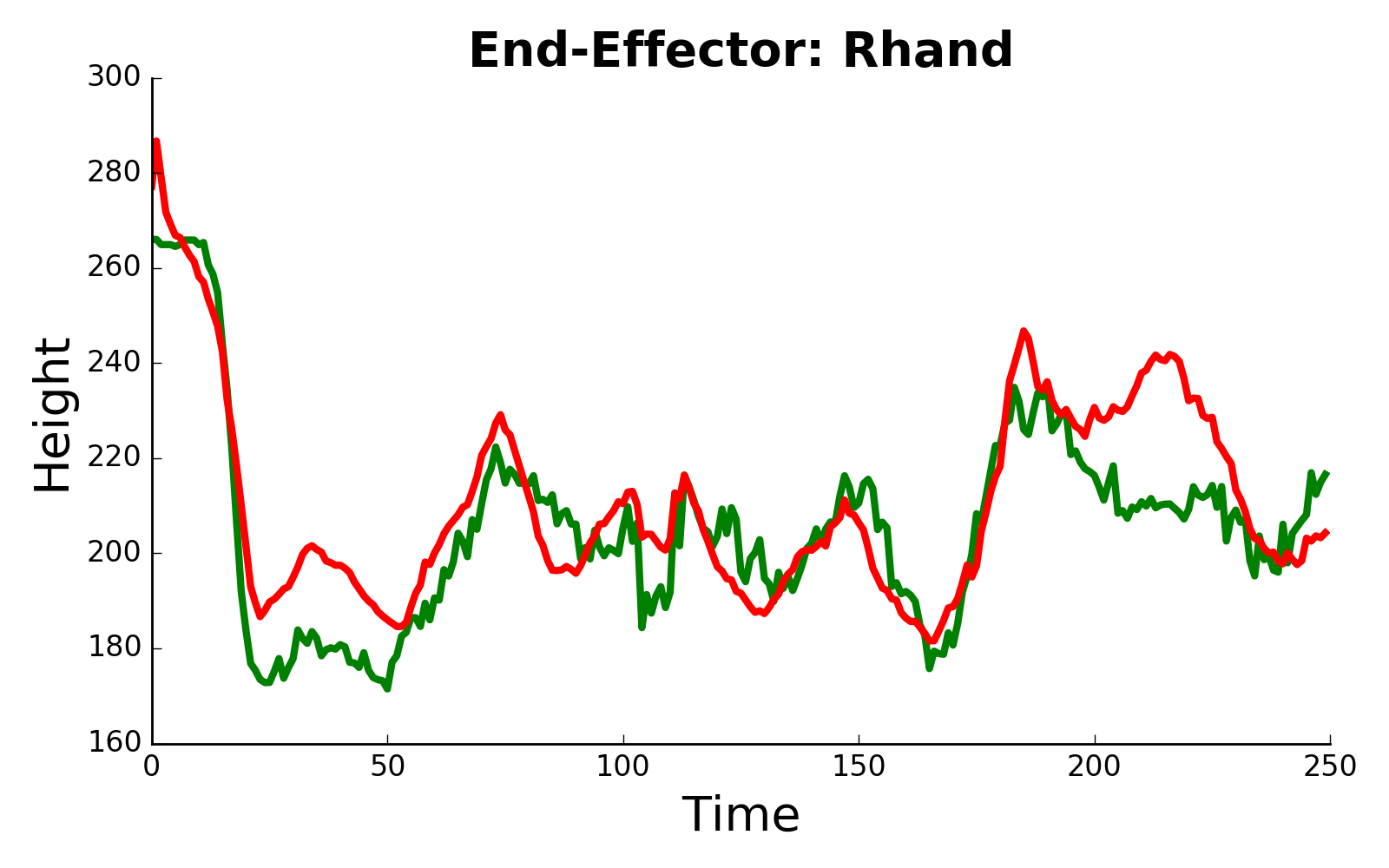} \hspace{-5pt}
	    \includegraphics[width=.25\linewidth]{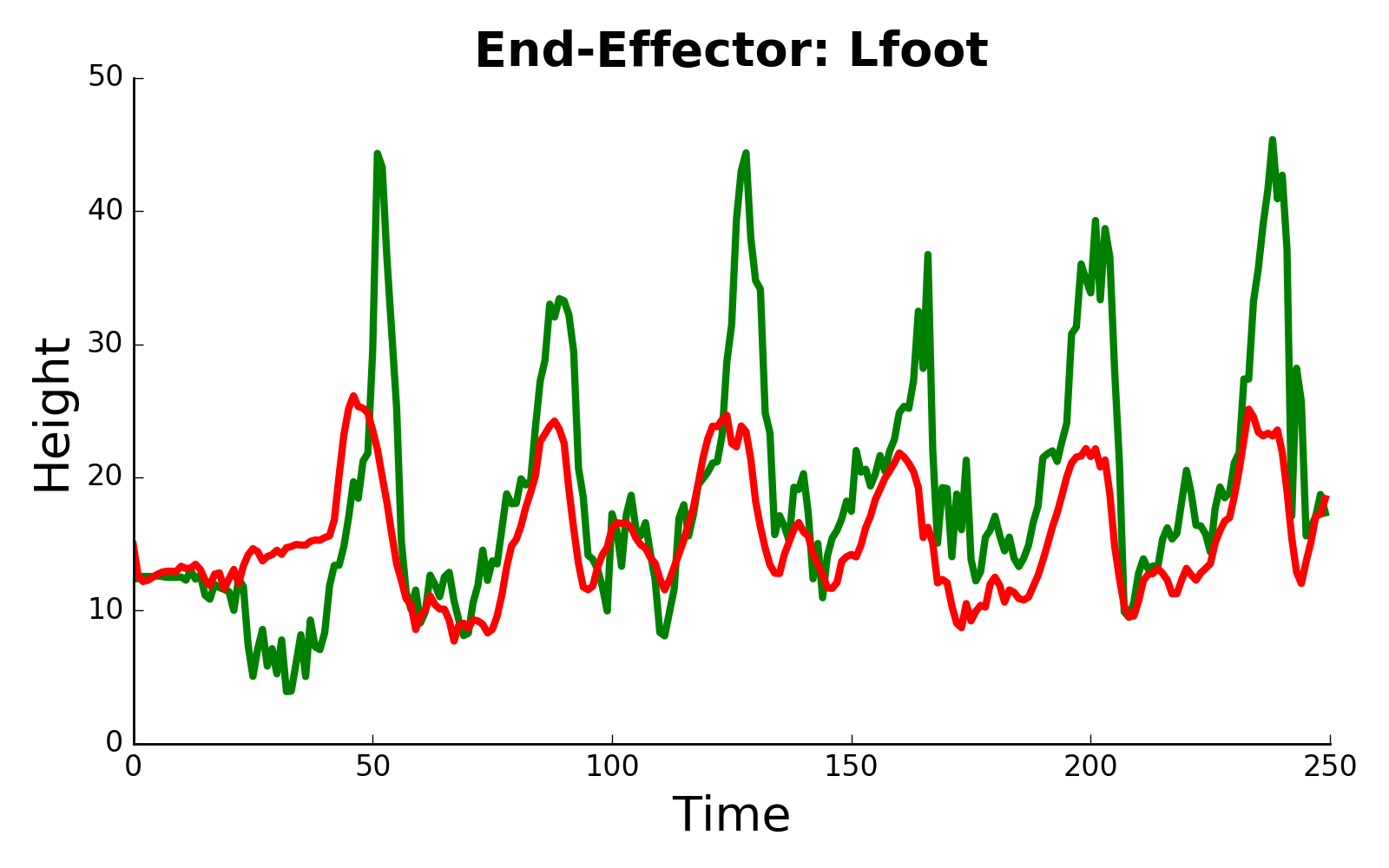} \hspace{-5pt}
	    \includegraphics[width=.25\linewidth]{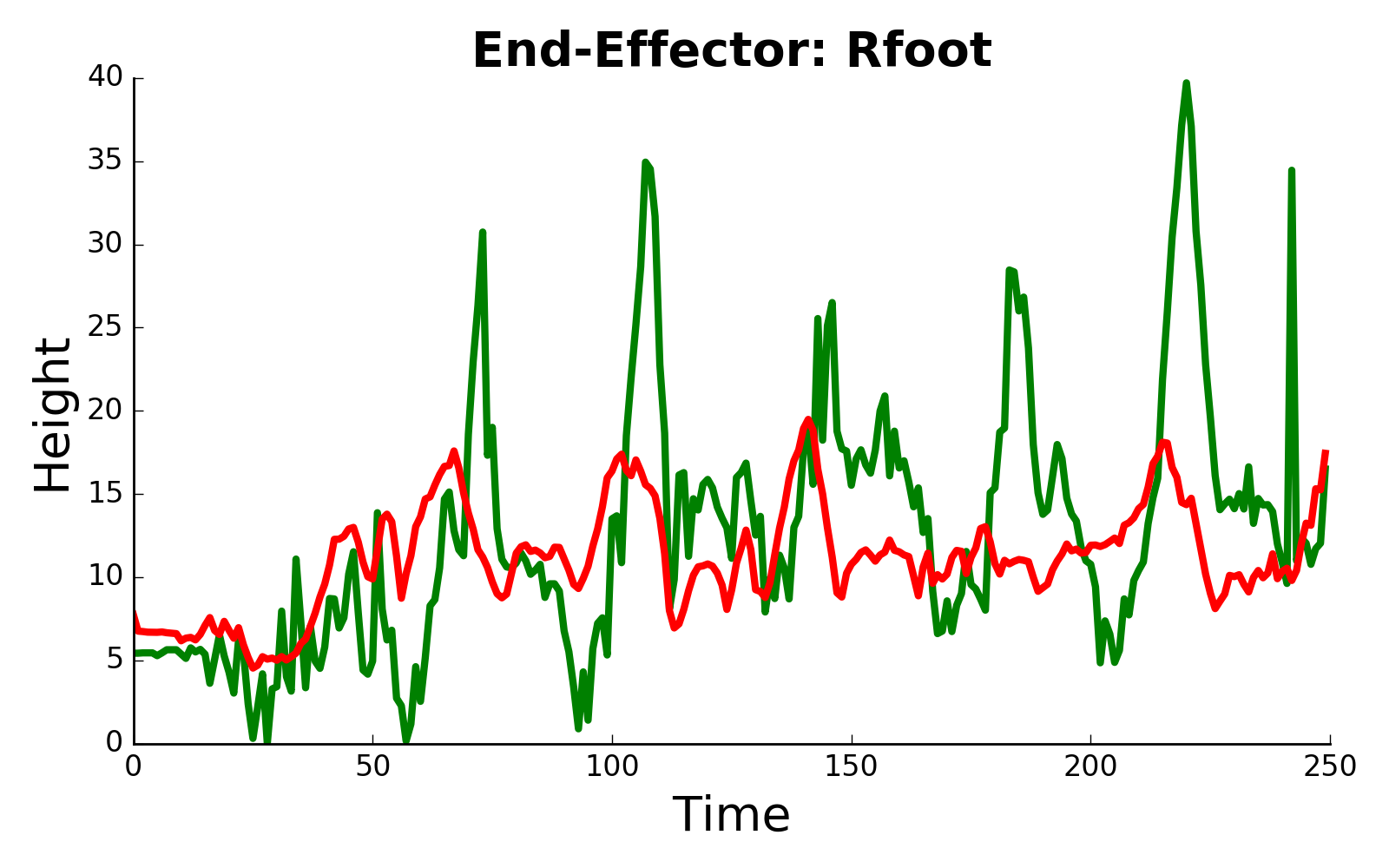} \hspace{-5pt} \\
	    
	    \includegraphics[width=.25\linewidth]{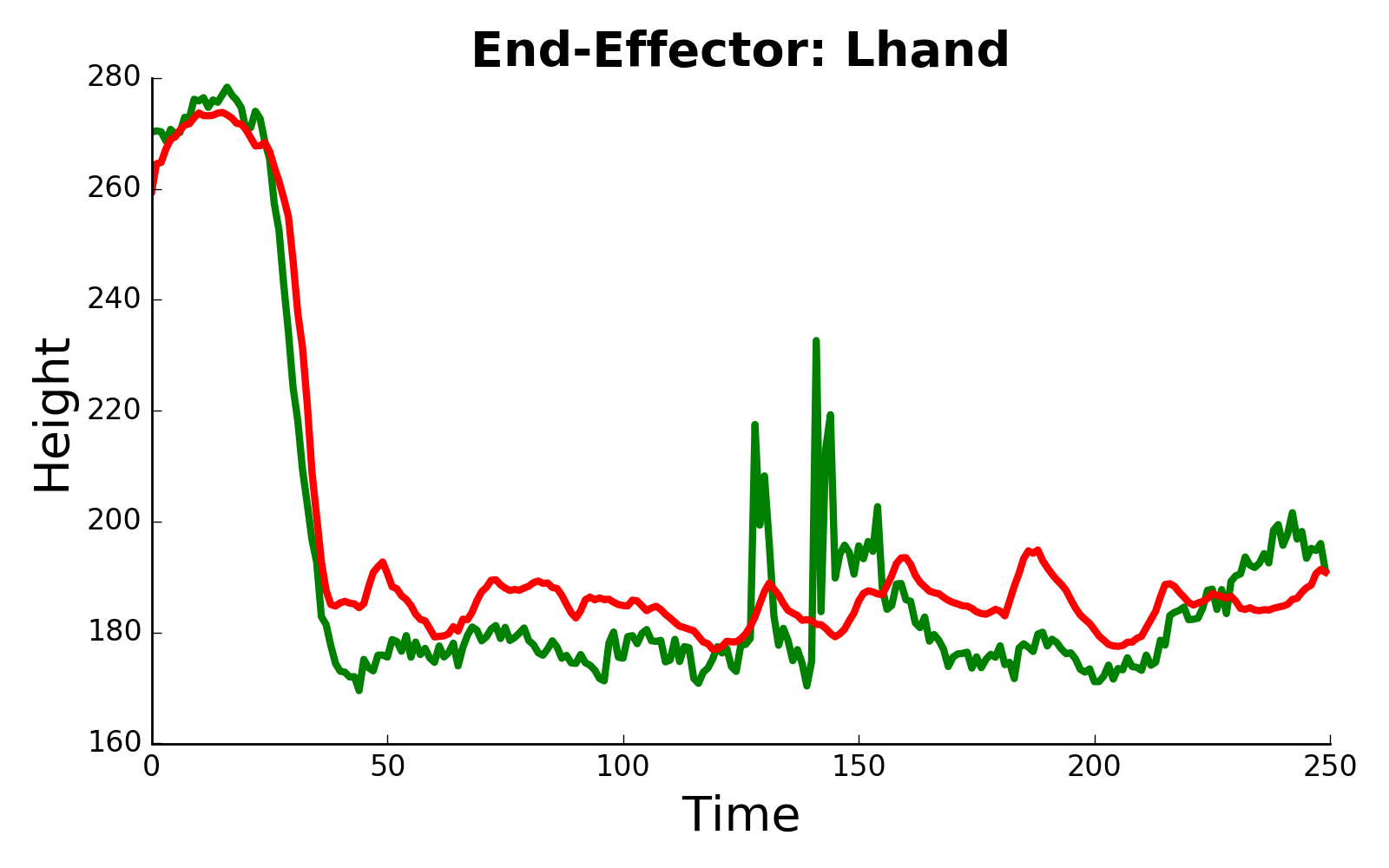} \hspace{-5pt}
	    \includegraphics[width=.25\linewidth]{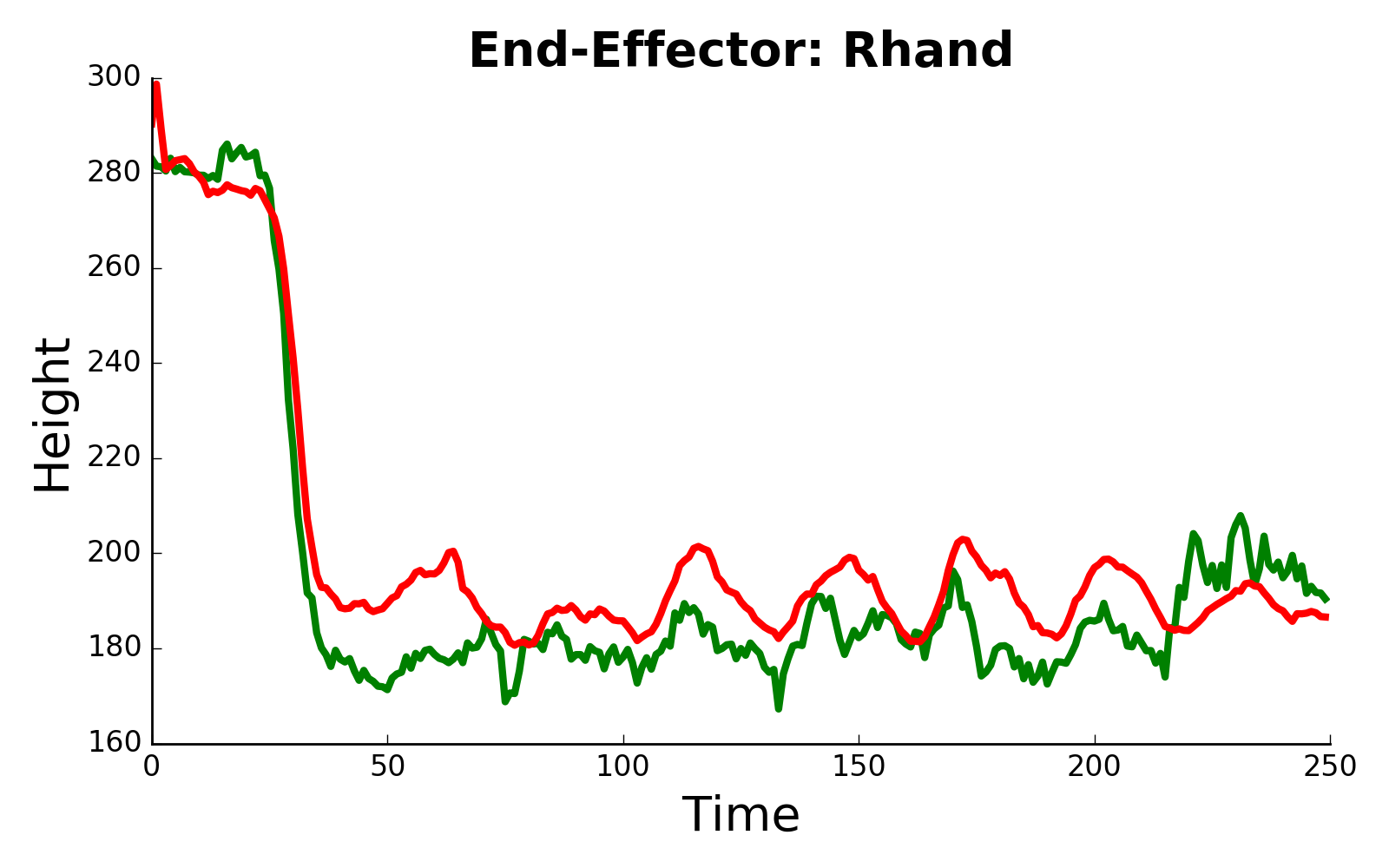} \hspace{-5pt}
	    \includegraphics[width=.25\linewidth]{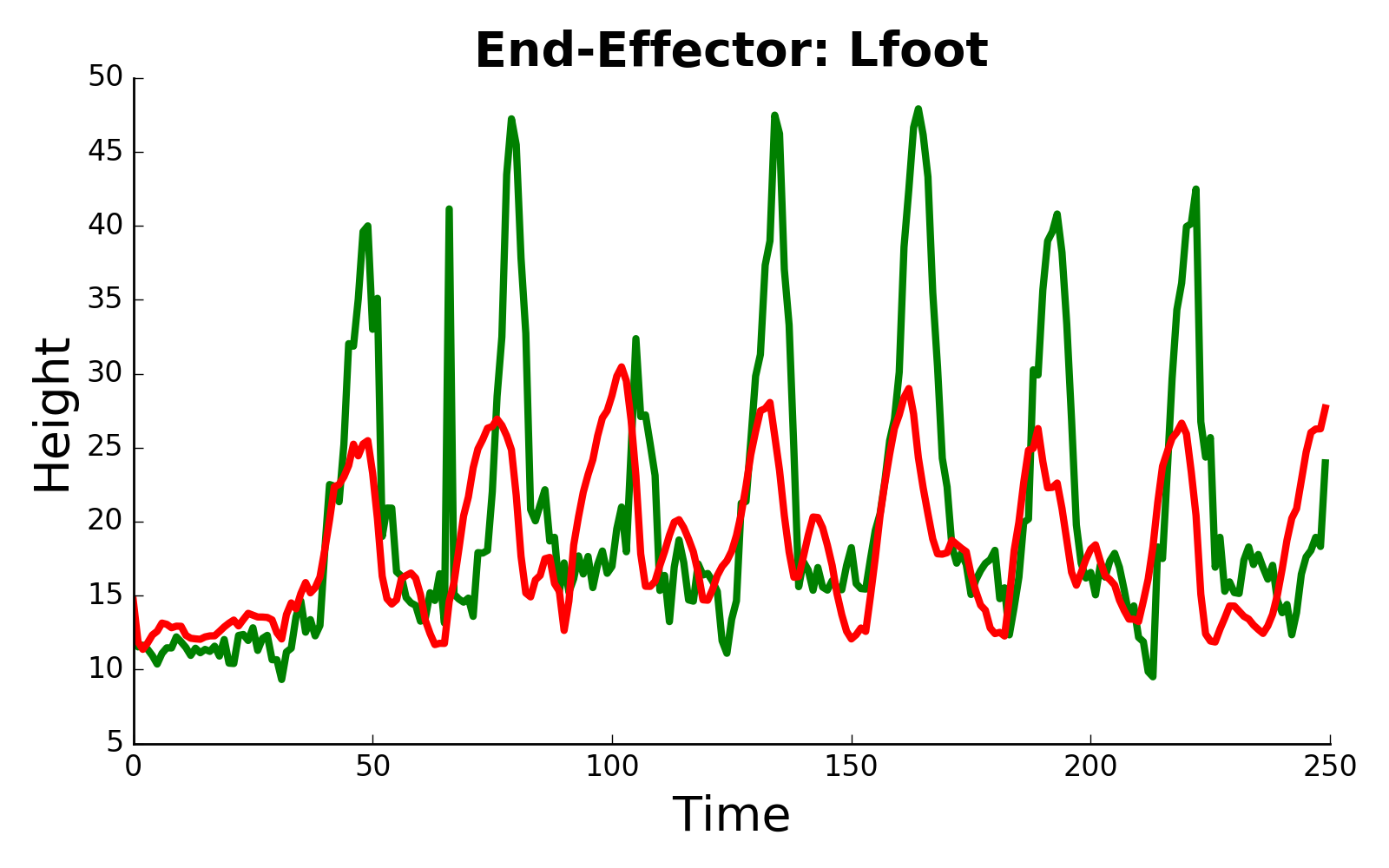} \hspace{-5pt}
	    \includegraphics[width=.25\linewidth]{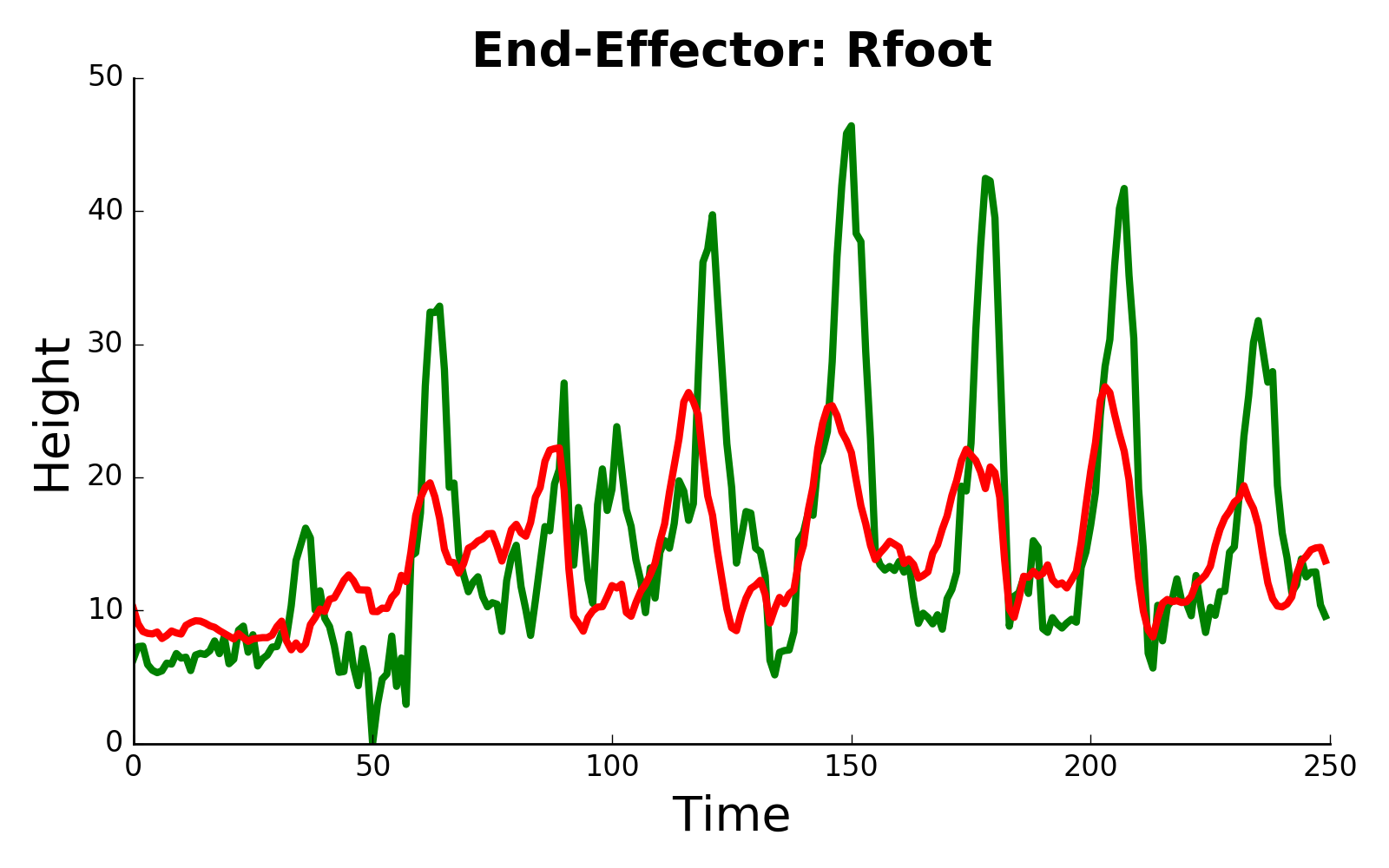} \hspace{-5pt} \\
	    
	    \includegraphics[width=.25\linewidth]{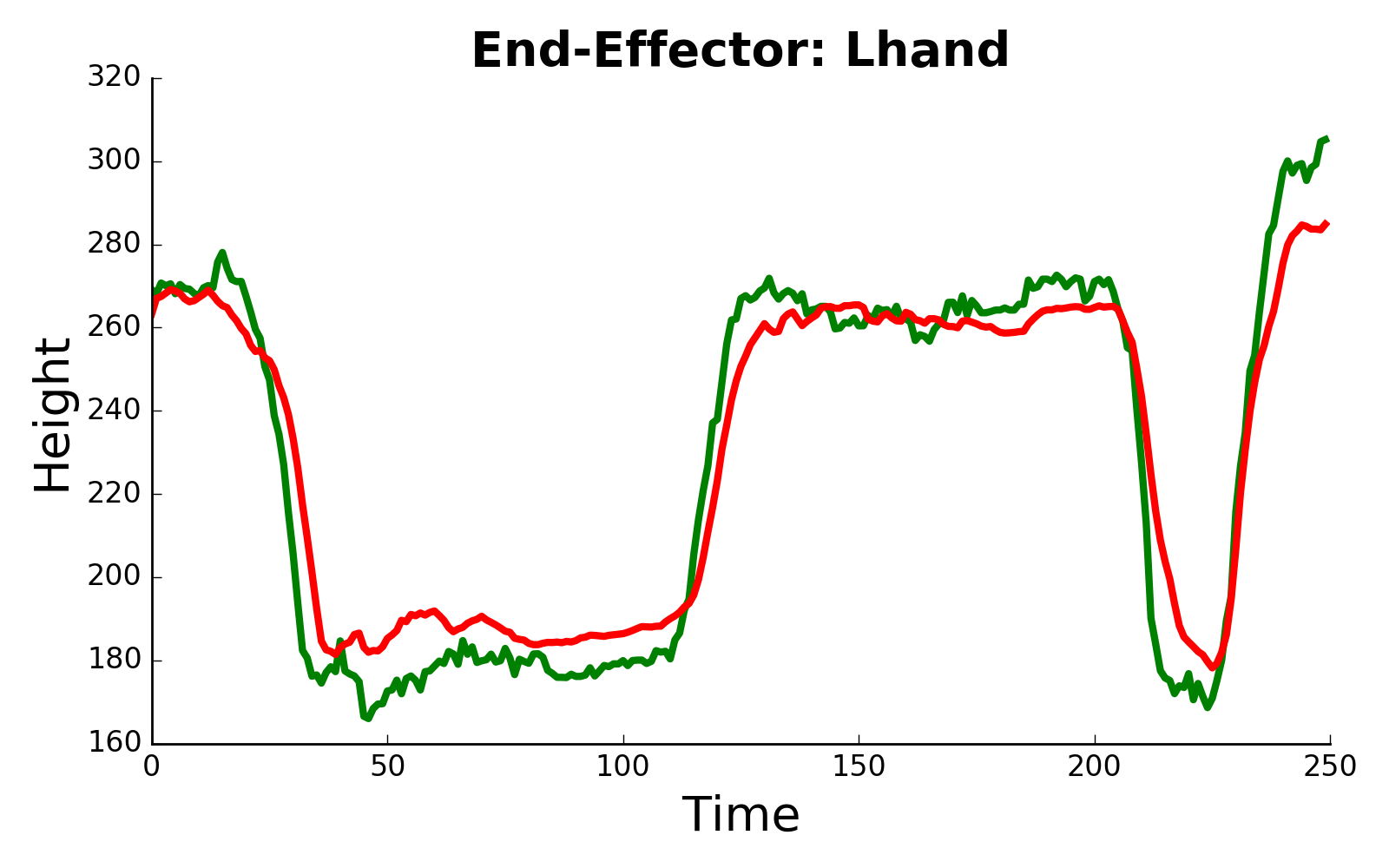} \hspace{-5pt}
	    \includegraphics[width=.25\linewidth]{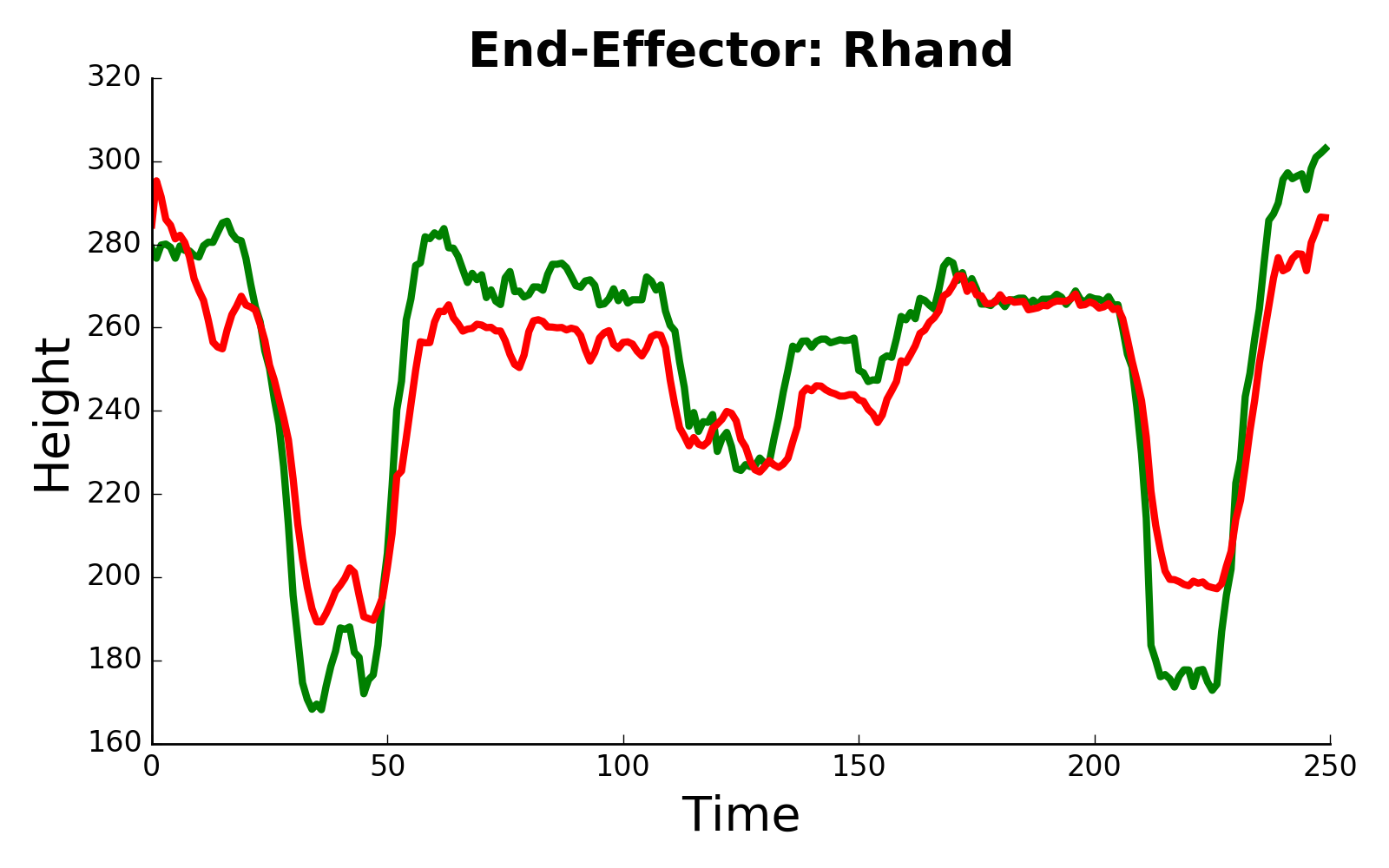} \hspace{-5pt}
	    \includegraphics[width=.25\linewidth]{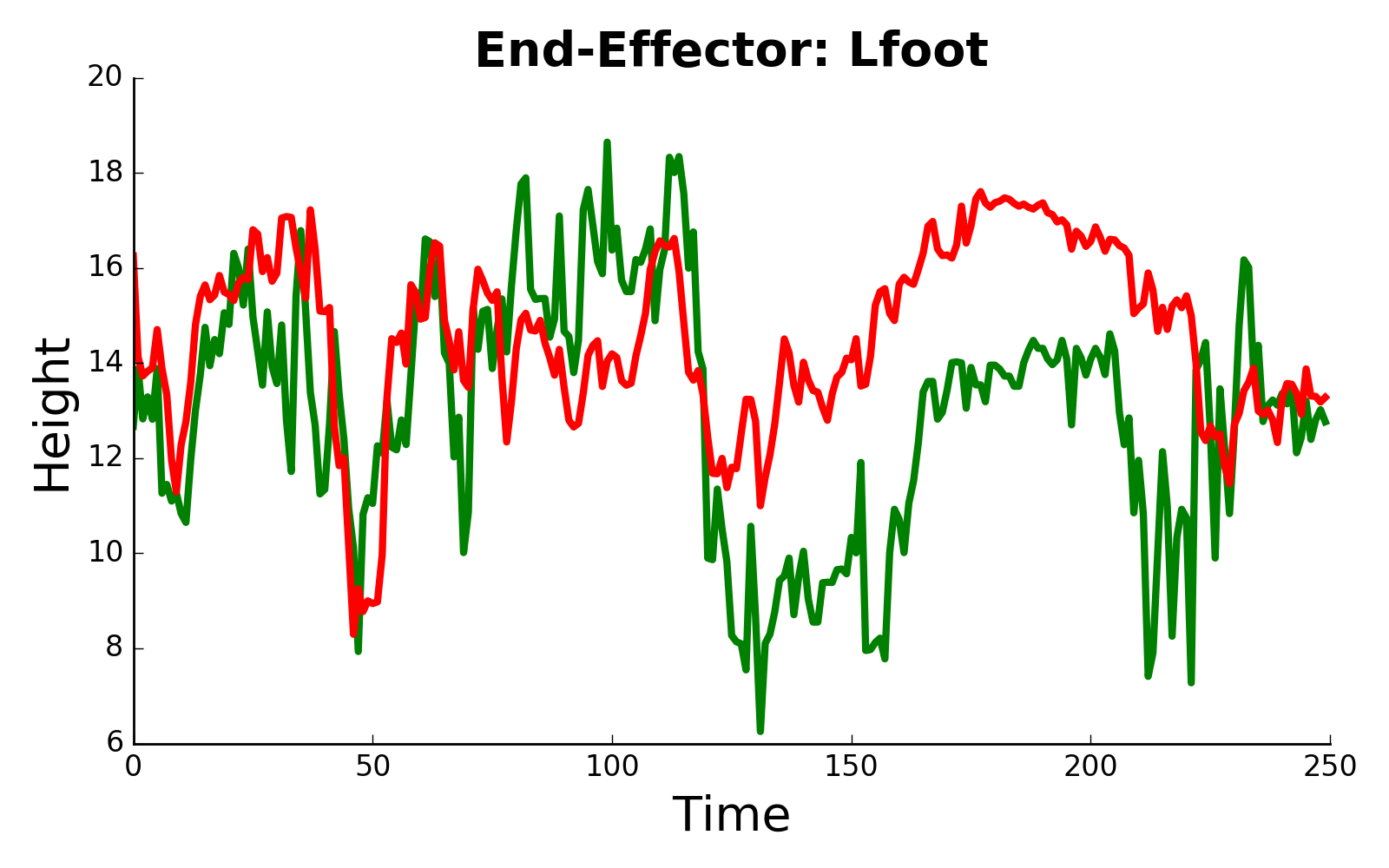} \hspace{-5pt}
	    \includegraphics[width=.25\linewidth]{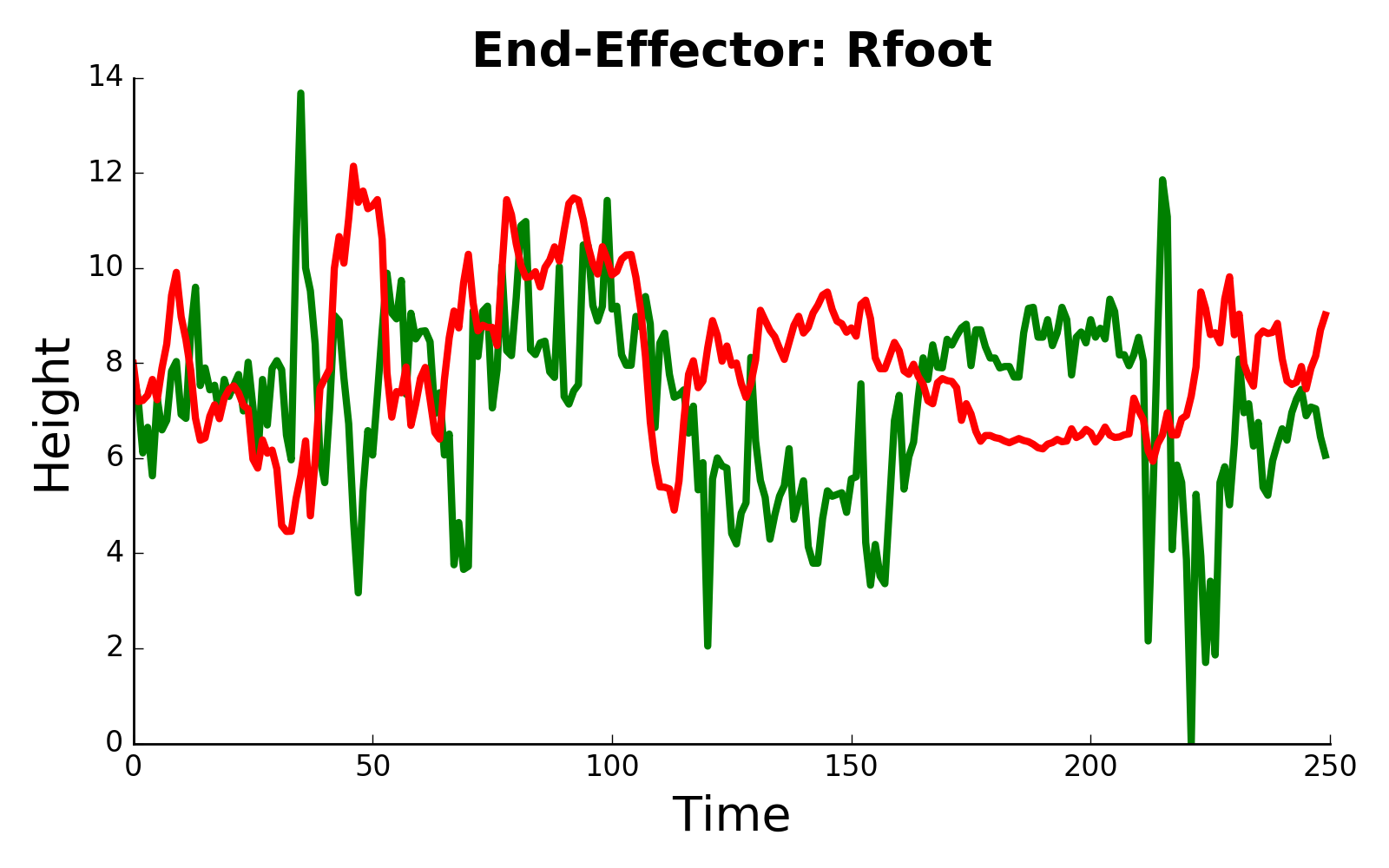} \hspace{-5pt} \\
	\end{subfigure}
	\end{center}
	\begin{center}
	\vspace{-.35in}
	\begin{subfigure}{0.06\linewidth}
	\includegraphics[width=1.0\linewidth]{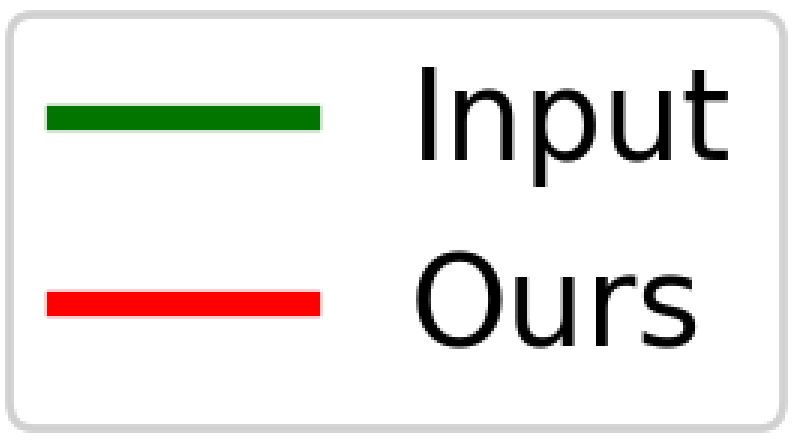}
	\end{subfigure}
	\vspace{-.2in}
	\end{center}
	\caption{3D pose estimation denoising. We present end-effector trajectoriess for 5 examples. Each row belongs a single example in the Human3.6M test set used in [16]. Please refer to our website for visual illustrations of the denoising results. \url{goo.gl/mDTvem}.}
	\vspace{-.1in}
\label{fig:h36m_plots}
\end{figure}

\noindent In Figure \ref{fig:h36m_plots}, we can see that our method denoises the hand end-effectors well, even without having trained with such data before.
The feet end-effector denoising is good as well, however in some cases it misses the overall feet height the original estimation had.
However, if we take a look at the provided videos, we can clearly see that our method's understanding of the input motion allows it to fix a lot of the shaking seen in the initially estimated 3D pose after motion retargetting.

\section{Demo Video and Qualitative Motion Retargetting Evaluation}
\noindent For the results demo video, please refer to the youtube video link \url{https://youtu.be/BGMyCFmGJWQ} (Note: The demo video contains audio. Please wear headphones if you believe you may disturb people around you).
For more videos, please go to \url{goo.gl/mDTvem}.

\section{Data collection process}
\noindent In this section, we describe the exact steps for collecting the training and testing data from the Mixamo website [1].
As training data, we collected 1656 unique motion sequences distributed over 7 different characters.
As testing data, we use 68 unique sequences of at least 4 seconds each (74 total) from which we extract 173 unique non-overlapping 4-second clips (185 total).
Please note that the last clip in each sequence which may overlap if there are less that 4 seconds left over after all non-overlapping clips have been extracted.
The Mixamo website contains motion separated by pages, the specific pages we downloaded for each character are specified in Table~\ref{data:train/test} below:

\vspace{15pt}
\begin{table}[h]
\centering
\begin{tabular}{lr}

%
\setlength{\tabcolsep}{3pt}
\begin{tabular}{c}
\Xhline{4\arrayrulewidth}
Training data \\
\begin{tabular}{l|c}
\Xhline{4\arrayrulewidth}
Character & Motion sequence page  \\
\Xhline{4\arrayrulewidth}
Malcolm & [1-5] \\
Warrok W Kurniawan & [6-10]  \\
Goblin D Shareyko & [11-15] \\
Kaya & [16-20]  \\
Peasant Man & [21-25]  \\
Big Vegas & [26-30] \\
AJ & [31-35]  \\
\Xhline{4\arrayrulewidth}
\end{tabular}
\end{tabular}
\vspace{-2pt}

&

\setlength{\tabcolsep}{3pt}
\begin{tabular}{c}
\Xhline{4\arrayrulewidth}
Test data \\
\begin{tabular}{l|c}
\Xhline{4\arrayrulewidth}
Input -> Target & Motion sequence page  \\
\Xhline{4\arrayrulewidth}
Malcolm & 28, 51  \\
Warrok W Kurniawan & 18, 52  \\
Liam & 23, 45  \\
Mutant & 33, 45, 52 \\
Claire & 52 \\
Sporty Granny & 51 \\
\ & \ \\
\Xhline{4\arrayrulewidth}
\end{tabular}
\end{tabular}
\end{tabular}
\caption{Data collection for each character and animation page in the Mixamo website.}
\label{data:train/test}
\end{table}

\noindent At test time, we perform motion retargetting for each testing scenario as shown in Tables~\ref{data:known_motion} and \ref{data:new_motion} below:

\vspace{5pt}
\begin{table}[ht]
\centering
\begin{tabular}{lr}

\setlength{\tabcolsep}{3pt}
\begin{tabular}{c}
\Xhline{4\arrayrulewidth}
Known Motion / Known Character \\
\begin{tabular}{l|c}
\Xhline{4\arrayrulewidth}
Input $\rightarrow$ Target & Motion sequence page  \\
\Xhline{4\arrayrulewidth}
Kaya $\rightarrow$ Warrok W Kurniawan & 18  \\
Big Vegas $\rightarrow$ Malcolm & 28  \\
\Xhline{4\arrayrulewidth}
\end{tabular}
\end{tabular}
&

\setlength{\tabcolsep}{3pt}
\begin{tabular}{c}
\Xhline{4\arrayrulewidth}
Known Motion / New Character \\
\begin{tabular}{l|c}
\Xhline{4\arrayrulewidth}
Input $\rightarrow$ Target & Motion sequence page  \\
\Xhline{4\arrayrulewidth}
Peasant Man $\rightarrow$ Liam & 23  \\
AJ $\rightarrow$ Mutant & 33  \\
\Xhline{4\arrayrulewidth}
\end{tabular}
\end{tabular}

\end{tabular}
\caption{Quantitative evaluation of online motion retargetting using mean square error (MSE). Case study: Known motion / known character (left), and known motion / new character (right).}
\label{data:known_motion}
\end{table}


\begin{table}[ht]
\centering
\begin{tabular}{lr}

\setlength{\tabcolsep}{3pt}
\begin{tabular}{c}
\Xhline{4\arrayrulewidth}
New Motion / Known Character \\
\begin{tabular}{l|c}
\Xhline{4\arrayrulewidth}
Input $\rightarrow$ Target & Motion sequence page  \\
\Xhline{4\arrayrulewidth}
Sporty Granny $\rightarrow$ Malcolm & 51  \\
Claire $\rightarrow$ Warrok W Kurniawan & 52  \\
\Xhline{4\arrayrulewidth}
\end{tabular}
\end{tabular}
\vspace{-2pt}

&

\setlength{\tabcolsep}{3pt}
\begin{tabular}{c}
\Xhline{4\arrayrulewidth}
New Motion / New Character \\
\begin{tabular}{l|c}
\Xhline{4\arrayrulewidth}
Input $\rightarrow$ Target & Motion sequence page  \\
\Xhline{4\arrayrulewidth}
Mutant $\rightarrow$ Liam & 45  \\
Claire $\rightarrow$ Mutant & 52  \\
\Xhline{4\arrayrulewidth}
\end{tabular}
\end{tabular}
\end{tabular}
\caption{Quantitative evaluation of online motion retargetting using mean square error (MSE). Case study: New motion / known character (left), and new motion / new character (right).}
\label{data:new_motion}
\end{table}

\vspace{15pt}

\newpage
\section{Architecture and training details}
In this section, we provide the network architectures detailes used throughout this paper.The RNN architectures are implemented by a 2-layer Gated Recurrent Unit (GRU) with 512-dimensional hidden state.
As the discriminator network, we use a 5 layer 1D fully-convolutional neural network with size 4 kernel, and convolutions across.
Layers 1-4 have leakyReLU activations with leak of 0.2, dropout of 0.7 keep probability, “same” convolution output with stride 2.
Layers 2-4 each have a instance normalization layer with default parameters in the tensorflow implementation.
The last layer implements a “valid” convolution with linear activation.
For training the networks, we use the Adam optimizer with a learning rate of 1e-4 for both the retargetting RNN and Discriminator, and clip the RNN gradients by global norm of 25.
We also implemented a balancing technique between the retargetting network and the discriminator, where the discriminator is not updated if the probability of the generator output being a real falls below $0.3$.

\end{appendix}

%% file: main.bbl
\begin{thebibliography}{10}\itemsep=-1pt

\bibitem{Mixamo}
{Adobe's Mixamo}.
\newblock \url{https://www.mixamo.com}.
\newblock Accessed: 2017-09-28.

\bibitem{Ayusawa2017TRO}
K.~Ayusawa and E.~Yoshida.
\newblock Motion retargeting for humanoid robots based on simultaneous morphing
  parameter identification and motion optimization.
\newblock {\em IEEE Trans. on Robotics}, 33(6), 2017.
\newblock to appear.

\bibitem{Bagnell}
J.~A.~D. Bagnell.
\newblock An invitation to imitation.
\newblock Technical Report CMU-RI-TR-15-08, Pittsburgh, PA, March 2015.

\bibitem{bin2015kinodynamically}
G.~Bin~Hammam, P.~M. Wensing, B.~Dariush, and D.~E. Orin.
\newblock Kinodynamically consistent motion retargeting for humanoids.
\newblock {\em International Journal of Humanoid Robotics}, 12(04):1550017,
  2015.

\bibitem{blender}
{Blender Online Community}.
\newblock {\em Blender - a 3D modelling and rendering package}.
\newblock Blender Foundation, Blender Institute, Amsterdam, 2017.

\bibitem{brand2000style}
M.~Brand and A.~Hertzmann.
\newblock Style machines.
\newblock In {\em Proceedings of the 27th annual conference on Computer
  graphics and interactive techniques}, pages 183--192. ACM
  Press/Addison-Wesley Publishing Co., 2000.

\bibitem{butepage2017deep}
J.~B{\"u}tepage, M.~J. Black, D.~Kragic, and H.~Kjellstr{\"o}m.
\newblock Deep representation learning for human motion prediction and
  classification.
\newblock In {\em IEEE Conference on Computer Vision and Pattern Recognition
  (CVPR)}, page 2017, 2017.

\bibitem{chang2016compositional}
M.~B. Chang, T.~Ullman, A.~Torralba, and J.~B. Tenenbaum.
\newblock A compositional object-based approach to learning physical dynamics.
\newblock In {\em ICLR}, 2017.

\bibitem{choi}
K.-J. Choi and H.-S. Ko.
\newblock On-line motion retargetting.
\newblock In {\em Computer Graphics and Applications, 1999. Proceedings.
  Seventh Pacific Conference on}, pages 32--42. IEEE, 1999.

\bibitem{fragkiadaki2015recurrent}
K.~Fragkiadaki, S.~Levine, P.~Felsen, and J.~Malik.
\newblock Recurrent network models for human dynamics.
\newblock In {\em Proceedings of the IEEE International Conference on Computer
  Vision}, pages 4346--4354, 2015.

\bibitem{gleicher}
M.~Gleicher.
\newblock Retargetting motion to new characters.
\newblock In {\em Proceedings of the 25th annual conference on Computer
  graphics and interactive techniques}, pages 33--42. ACM, 1998.

\bibitem{NIPS2014_5423}
I.~Goodfellow, J.~Pouget-Abadie, M.~Mirza, B.~Xu, D.~Warde-Farley, S.~Ozair,
  A.~Courville, and Y.~Bengio.
\newblock Generative adversarial nets.
\newblock In {\em Advances in neural information processing systems}, pages
  2672--2680, 2014.

\bibitem{graves2013speech}
A.~Graves, A.-r. Mohamed, and G.~Hinton.
\newblock Speech recognition with deep recurrent neural networks.
\newblock In {\em Acoustics, speech and signal processing (icassp), 2013 ieee
  international conference on}, pages 6645--6649. IEEE, 2013.

\bibitem{grochow2004style}
K.~Grochow, S.~L. Martin, A.~Hertzmann, and Z.~Popovi{\'c}.
\newblock Style-based inverse kinematics.
\newblock In {\em ACM transactions on graphics (TOG)}, volume~23, pages
  522--531. ACM, 2004.

\bibitem{ho2016generative}
J.~Ho and S.~Ermon.
\newblock Generative adversarial imitation learning.
\newblock In {\em Advances in Neural Information Processing Systems}, pages
  4565--4573, 2016.

\bibitem{Holden}
D.~Holden, J.~Saito, and T.~Komura.
\newblock A deep learning framework for character motion synthesis and editing.
\newblock {\em ACM Transactions on Graphics (TOG)}, 35(4):138, 2016.

\bibitem{hsu2005style}
E.~Hsu, K.~Pulli, and J.~Popovi{\'c}.
\newblock Style translation for human motion.
\newblock In {\em ACM Transactions on Graphics (TOG)}, volume~24, pages
  1082--1089. ACM, 2005.

\bibitem{h36m_pami}
C.~Ionescu, D.~Papava, V.~Olaru, and C.~Sminchisescu.
\newblock {Human3.6M}: Large scale datasets and predictive methods for {3D}
  human sensing in natural environments.
\newblock {\em PAMI}, 36(7):1325--1339, 2014.

\bibitem{jain2016structural}
A.~Jain, A.~R. Zamir, S.~Savarese, and A.~Saxena.
\newblock Structural-rnn: Deep learning on spatio-temporal graphs.
\newblock In {\em Proceedings of the IEEE Conference on Computer Vision and
  Pattern Recognition}, pages 5308--5317, 2016.

\bibitem{johnson2016google}
M.~Johnson, M.~Schuster, Q.~V. Le, M.~Krikun, Y.~Wu, Z.~Chen, N.~Thorat,
  F.~Vi{\'e}gas, M.~Wattenberg, G.~Corrado, M.~Hughes, and J.~Dean.
\newblock Google's multilingual neural machine translation system: Enabling
  zero-shot translation.
\newblock {\em Transactions of the Association of Computational Linguistics},
  5:339--351, 2017.

\bibitem{kovar2004automated}
L.~Kovar and M.~Gleicher.
\newblock Automated extraction and parameterization of motions in large data
  sets.
\newblock In {\em ACM Transactions on Graphics (ToG)}, volume~23, pages
  559--568. ACM, 2004.

\bibitem{lee1999hierarchical}
J.~Lee and S.~Y. Shin.
\newblock A hierarchical approach to interactive motion editing for human-like
  figures.
\newblock In {\em Proceedings of the 26th annual conference on Computer
  graphics and interactive techniques}, pages 39--48. ACM Press/Addison-Wesley
  Publishing Co., 1999.

\bibitem{li2017auto}
Z.~Li, Y.~Zhou, S.~Xiao, C.~He, and H.~Li.
\newblock Auto-conditioned lstm network for extended complex human motion
  synthesis.
\newblock In {\em ICLR}, 2018.

\bibitem{liu2017material}
G.~Liu, D.~Ceylan, E.~Yumer, J.~Yang, and J.-M. Lien.
\newblock Material editing using a physically based rendering network.
\newblock In {\em 2017 IEEE International Conference on Computer Vision
  (ICCV)}, pages 2280--2288. IEEE, 2017.

\bibitem{martinez2017human}
J.~Martinez, M.~J. Black, and J.~Romero.
\newblock On human motion prediction using recurrent neural networks.
\newblock In {\em 2017 IEEE Conference on Computer Vision and Pattern
  Recognition (CVPR)}, pages 4674--4683. IEEE, 2017.

\bibitem{martinez2017}
J.~Martinez, R.~Hossain, J.~Romero, and J.~J. Little.
\newblock A simple yet effective baseline for 3d human pose.
\newblock {\em ICCV}, 2017.

\bibitem{mehta2017vnect}
D.~Mehta, S.~Sridhar, O.~Sotnychenko, H.~Rhodin, M.~Shafiei, H.-P. Seidel,
  W.~Xu, D.~Casas, and C.~Theobalt.
\newblock Vnect: Real-time 3d human pose estimation with a single rgb camera.
\newblock {\em ACM Transactions on Graphics (TOG)}, 36(4):44, 2017.

\bibitem{merel2017learning}
J.~Merel, Y.~Tassa, S.~Srinivasan, J.~Lemmon, Z.~Wang, G.~Wayne, and N.~Heess.
\newblock Learning human behaviors from motion capture by adversarial
  imitation.
\newblock {\em arXiv preprint arXiv:1707.02201}, 2017.

\bibitem{min2010synthesis}
J.~Min, H.~Liu, and J.~Chai.
\newblock Synthesis and editing of personalized stylistic human motion.
\newblock In {\em Proceedings of the 2010 ACM SIGGRAPH symposium on Interactive
  3D Graphics and Games}, pages 39--46. ACM, 2010.

\bibitem{rezende2016unsupervised}
D.~J. Rezende, S.~A. Eslami, S.~Mohamed, P.~Battaglia, M.~Jaderberg, and
  N.~Heess.
\newblock Unsupervised learning of 3d structure from images.
\newblock In {\em Advances In Neural Information Processing Systems}, pages
  4996--5004, 2016.

\bibitem{rose2001artist}
C.~F. Rose~III, P.-P.~J. Sloan, and M.~F. Cohen.
\newblock Artist-directed inverse-kinematics using radial basis function
  interpolation.
\newblock In {\em Computer Graphics Forum}, volume~20, pages 239--250. Wiley
  Online Library, 2001.

\bibitem{schaal1999imitation}
S.~Schaal.
\newblock Is imitation learning the route to humanoid robots?
\newblock {\em Trends in cognitive sciences}, 3(6):233--242, 1999.

\bibitem{TCN2017}
P.~Sermanet, C.~Lynch, J.~Hsu, and S.~Levine.
\newblock Time-contrastive networks: Self-supervised learning from multi-view
  observation.
\newblock {\em arXiv preprint arXiv:1704.06888}, 2017.

\bibitem{Shon2006Imitation}
A.~Shon, K.~Grochow, A.~Hertzmann, and R.~P. Rao.
\newblock Learning shared latent structure for image synthesis and robotic
  imitation.
\newblock In Y.~Weiss, P.~B. Sch\"{o}lkopf, and J.~C. Platt, editors, {\em
  Advances in Neural Information Processing Systems 18}, pages 1233--1240. MIT
  Press, 2006.

\bibitem{tak2005physically}
S.~Tak and H.-S. Ko.
\newblock A physically-based motion retargeting filter.
\newblock {\em ACM Transactions on Graphics (TOG)}, 24(1):98--117, 2005.

\bibitem{taylor2007modeling}
G.~W. Taylor, G.~E. Hinton, and S.~T. Roweis.
\newblock Modeling human motion using binary latent variables.
\newblock In {\em Advances in neural information processing systems}, pages
  1345--1352, 2007.

\bibitem{tulsiani2017multi}
S.~Tulsiani, T.~Zhou, A.~A. Efros, and J.~Malik.
\newblock Multi-view supervision for single-view reconstruction via
  differentiable ray consistency.
\newblock In {\em CVPR}, volume~1, page~3, 2017.

\bibitem{wang2008gaussian}
J.~M. Wang, D.~J. Fleet, and A.~Hertzmann.
\newblock Gaussian process dynamical models for human motion.
\newblock {\em IEEE transactions on pattern analysis and machine intelligence},
  30(2):283--298, 2008.

\bibitem{vda}
J.~Wu, E.~Lu, P.~Kohli, W.~T. Freeman, and J.~B. Tenenbaum.
\newblock Learning to see physics via visual de-animation.
\newblock In {\em Advances in Neural Information Processing Systems}, 2017.

\bibitem{xia2015realtime}
S.~Xia, C.~Wang, J.~Chai, and J.~Hodgins.
\newblock Realtime style transfer for unlabeled heterogeneous human motion.
\newblock {\em ACM Transactions on Graphics (TOG)}, 34(4):119, 2015.

\bibitem{yan2016perspective}
X.~Yan, J.~Yang, E.~Yumer, Y.~Guo, and H.~Lee.
\newblock Perspective transformer nets: Learning single-view 3d object
  reconstruction without 3d supervision.
\newblock In {\em Advances in Neural Information Processing Systems}, pages
  1696--1704, 2016.

\bibitem{yumer2016spectral}
M.~E. Yumer and N.~J. Mitra.
\newblock Spectral style transfer for human motion between independent actions.
\newblock {\em ACM Transactions on Graphics (TOG)}, 35(4):137, 2016.

\bibitem{zhou2016learning}
T.~Zhou, P.~Krahenbuhl, M.~Aubry, Q.~Huang, and A.~A. Efros.
\newblock Learning dense correspondence via 3d-guided cycle consistency.
\newblock In {\em Proceedings of the IEEE Conference on Computer Vision and
  Pattern Recognition}, pages 117--126, 2016.

\bibitem{CycleGAN2017}
J.-Y. Zhu, T.~Park, P.~Isola, and A.~A. Efros.
\newblock Unpaired image-to-image translation using cycle-consistent
  adversarial networks.
\newblock In {\em ICCV}, 2017.

\end{thebibliography}
